\documentclass[11pt,a4paper]{article}

\usepackage[T1]{fontenc}
\usepackage[utf8]{inputenc}
\usepackage{lmodern}
\usepackage{microtype}
\usepackage[margin=1in]{geometry}
\usepackage{amsmath,amssymb,amsthm,mathtools}
\usepackage{booktabs}
\usepackage{array}
\usepackage{longtable}
\usepackage{tabularx}
\usepackage{multirow}
\usepackage{makecell}
\usepackage{enumitem}
\usepackage{xcolor}
\usepackage[colorlinks=true,linkcolor=blue!50!black,urlcolor=blue!60!black,citecolor=blue!60!black]{hyperref}
\usepackage{caption}
\usepackage{url}
\usepackage{fancyvrb}
\usepackage{listings}
\usepackage{graphicx}
\usepackage{adjustbox}
\usepackage{wrapfig}
\usepackage{float}

\setlist[itemize]{leftmargin=1.4em,topsep=2pt,parsep=2pt,itemsep=2pt}
\setlist[enumerate]{leftmargin=1.6em,topsep=2pt,parsep=2pt,itemsep=2pt}

\usepackage{titlesec}
\titlespacing*{\section}{0pt}{1.6\baselineskip}{0.6\baselineskip}
\titlespacing*{\subsection}{0pt}{1.1\baselineskip}{0.4\baselineskip}
\titlespacing*{\subsubsection}{0pt}{0.9\baselineskip}{0.3\baselineskip}

\theoremstyle{plain}

\theoremstyle{definition}

\renewcommand{\Re}{\operatorname{Re}}
\renewcommand{\Im}{\operatorname{Im}}

\title{\textbf{Complex-Valued Phase-Coherent Transformer}}

\author{Leona Hioki\\
\texttt{leohio@pm.me}}

\date{}

\begin{document}
\maketitle

\begin{center}
\begin{minipage}{0.9\textwidth}
\color{red}
\textbf{This version has been \emph{deprecated}. Please refer to the new version at \url{https://arxiv.org/pdf/2609.22415}. A retraction of this paper was not necessary; however, the new version provides an updated conclusion that integrates new experimental results.}
\end{minipage}
\end{center}
\vspace{0.5\baselineskip}

\small

\begin{abstract}
\setlength{\parindent}{0pt}
\setlength{\parskip}{0.6\baselineskip}
\noindent\ignorespaces
Complex-valued Transformers have largely inherited softmax attention from real-valued architectures. However, row-normalised token competition is not necessarily aligned with phase-preserving computation. In this paper, we introduce the Phase-Coherent Transformer (PCT), which applies a real-valued, element-independent, smooth gate to L2-normalised complex query-key similarities. PCT replaces token competition with token-non-competing attention and is designed to preserve phase information across layers.

Across mid-scale benchmarks spanning long-range memory, hierarchical long-range reasoning, positional retrieval, phase-based memory and superposition, and image classification, PCT shows strong generalisation across task categories. Under parameter-fair comparison, PCT consistently outperforms both the standard softmax Transformer and its direct complex-valued counterpart. Moreover, even on tasks traditionally considered difficult for complex-valued neural networks, such as NIAH and LRA-Text, PCT remains competitive with Multiscreen, the strongest real-valued NN baseline in our comparison. Experiments introducing gates that deliberately violate the PCT conditions show that the design is not incidental: smooth gates that preserve negatively aligned phase components remain strong, whereas gates that delete such components collapse on long-range retrieval, and gates whose outputs become excessively large suffer clear performance degradation. PCT also shows no depth-related accuracy collapse across the tested depth range. These results support introducing multi-layer phase-coherent structure into attention as a promising design principle for achieving generalisation in complex-valued Transformers.

\end{abstract}

\medskip

\section{Introduction}
Complex-valued neural networks (CvNNs) have long not been viewed as candidates for general-purpose architectures. The literature that has developed since Trabelsi 2017 has centred on signal-processing domains where the input is intrinsically represented as complex numbers -- PolSAR classification, wireless communication, SAR target recognition, MRI, and audio spectrogram analysis. The implicit understanding has been that CvNNs are strong on complex-valued signals but cannot match real-valued NNs on general-purpose domains such as text, vision, and symbolic reasoning.

Behind this lies an underdevelopment of the architectural research itself. While complex extensions of convolutions, normalisation layers, and activations have been studied systematically, the Transformer side -- including well-designed enhancements oriented towards image tasks such as Eilers $\mathbb{C}$Att and Holographic Transformer -- has not seriously explored anything beyond softmax attention. Attention mechanisms genuinely fitted to a complex-valued substrate -- primitives that depart substantially from softmax, which was optimised for real-valued NNs -- have not been systematically investigated.

Looking at the situation from a different angle, two observations point in the same direction:

\begin{enumerate}
  \item \textbf{On the real-valued NN side}, Multiscreen (Nakanishi 2026, \emph{Screening Is Enough}) has shown that an attention primitive can substantially outperform softmax Transformers on long-range dependency tasks, undermining the assumption that softmax attention is universally optimal. Room to fundamentally redesign the attention primitive has opened up even for real-valued NNs.
  \item \textbf{On the complex-valued NN side specifically}, complex weights and inputs carry phase information, and a deep network can perform meaningful computation only if phase information is preserved across layers. There is no a priori reason to expect that attention structures developed for real-valued NNs -- and the row normalisation of softmax in particular -- would contribute to this multi-layer phase preservation. On the contrary, softmax row normalisation structurally introduces a "token competition" in which tokens compete for attention mass, and this acts as an unnecessary constraint on a complex computation whose layers must preserve phase.
\end{enumerate}

Combining these two observations, a natural conclusion follows: what should be tried first in a complex NN is \textbf{token non-competition} -- an attention primitive whose weights are not normalised to sum to one within a row. This paper formalises the idea as the \textbf{Phase-Coherent Transformer (PCT)}: a sigmoid gate on the cosine score of L2-normalised complex Q, K used as the attention weight, with no row normalisation whatsoever. It is the most natural lift of sigmoid attention (Ramapuram 2024) to complex space, and forms a single family together with neighbouring variants such as tanh+1, softplus, and the complex version of screening attention (Nakanishi 2026).

We carry out a systematic 6-cell comparison on 9 mid-scale benchmarks and show that the PCT family consistently outperforms both existing real- and complex-valued Transformers under parameter-fair conditions ($real\_dim = complex\_dim \times 1.41$), achieves 1.000 on long-range NIAH -- a task on which CvNNs have traditionally been considered weak -- and exhibits no depth collapse over the depth 2--20 scaling range, addressing a long-standing concern about depth scalability in CvNNs.

\textbf{Results}:

\begin{enumerate}
  \item PCT and its neighbouring cells generalise across multiple tasks (R1).
  \item PCT retains dominance on NIAH, where complex Q, K offers no inductive advantage by construction (R2).
  \item PCT is uniquely LR/batch-robust on long-range Copy (R3).
  \item PCT outperforms the vanilla complex transformer (R4), the vanilla real transformer (R5), and the strongest real-side non-softmax baseline \texttt{real\_screen} (R6) under parameter-fair conditions.
  \item A mathematical interpretation -- token non-competition \& multi-layer phase preservation -- is given in \S{}5, with full proofs in Appendix M.
\end{enumerate}

In short, we experimentally observe strong signs of generalisation in complex-valued neural networks through the introduction of token-non-competing attention and multi-layer phase preservation, and analyzed the conditions of the signs with experiments described in \S{}5.

\medskip

\section{Related Work}
\subsection{Complex-valued neural networks}
Complex-valued neural networks have been studied as a way to model data whose natural representation contains both amplitude and phase. A major modern starting point is \emph{Deep Complex Networks} (Trabelsi et al., 2018), which introduced practical building blocks for complex-valued deep learning, including complex convolution, complex batch normalisation, complex initialisation, and end-to-end training recipes.\footnote{\url{https://www.microsoft.com/en-us/research/publication/deep-complex-networks/}} The underlying motivation was that complex-valued representations may offer richer representational capacity and more robust memory mechanisms, but had remained underused because standard neural-network components were missing.

Subsequent work has applied complex-valued models mainly to domains where phase is physically or mathematically meaningful, such as speech, audio, MRI, remote sensing, radar, wireless communication, and Fourier-domain signal processing. Software efforts such as \texttt{torchcvnn} have further consolidated this direction by providing reusable PyTorch components and datasets for complex-valued neural networks.\footnote{\url{https://github.com/torchcvnn/torchcvnn}} The practical success of complex-valued neural networks has therefore remained strongest in signal-domain or Fourier-domain settings, while their role as general-purpose sequence or reasoning architectures is much less developed.

This paper approaches that gap from the attention side. Instead of asking whether existing real-valued architectures can simply be ported into complex space, we ask whether the attention primitive itself should be redesigned for a complex-valued substrate.

\subsection{Complex-valued Transformers and complex attention}
A smaller line of work has adapted Transformer components to complex-valued representations. Eilers and Jiang (2023) introduced building blocks for a complex-valued Transformer architecture, including complex-valued scaled dot-product attention and complex-valued layer normalisation.\footnote{\url{https://sigport.org/documents/building-blocks-complex-valued-transformer-architecture}} Their experiments on MusicNet classification and sequence generation showed improved robustness to overfitting while maintaining performance comparable to real-valued Transformer baselines.

This line of work establishes that complex-valued Transformers are feasible. However, most existing complex Transformer designs still inherit the central structure of real-valued softmax attention: query-key scores are converted into row-normalised attention weights, and tokens compete for attention mass within each row. PCT differs at precisely this point. It does not merely replace real-valued layers by complex-valued layers; it changes the attention primitive itself by replacing row-normalised token competition with an element-independent real gate on complex query-key similarity.

\subsection{Alternatives to softmax attention}
The standard Transformer introduced scaled dot-product attention with row-wise softmax normalisation, which has become the default attention primitive for modern sequence models (Vaswani et al., 2017).\footnote{\url{https://arxiv.org/abs/1706.03762}} A large literature has since explored alternatives to softmax attention, often motivated by efficiency, stability, or scaling. Performer, for example, approximates softmax attention using positive orthogonal random features, achieving linear time and space complexity while retaining theoretical approximation guarantees (Choromanski et al., 2021).\footnote{\url{https://research.google/pubs/rethinking-attention-with-performers/}}

Other work replaces the softmax activation more directly. Sigmoid attention has recently been revisited with both theoretical and empirical analysis: Ramapuram et al. (2025) show that properly normalised sigmoid attention can act as a drop-in replacement for softmax across language, vision, and speech, while improving regularity and enabling efficient implementations.\footnote{\url{https://machinelearning.apple.com/research/sigmoid-self-attention}} We also include Multiscreen (Nakanishi 2026, \emph{Screening Is Enough}) as a recent threshold-based non-softmax attention primitive, using its real version as our strongest real-side baseline and its complex lift as a near-PCT comparison point.

These real-valued alternatives show that softmax is not the only viable attention primitive. However, their motivations are usually real-domain concerns: computational efficiency, regularity, stability, or approximation of softmax-like kernels. PCT is motivated by a different question: in a complex-valued substrate, where the value path carries phase information, which attention primitive is structurally compatible with preserving phase information across layers?

\subsection{Gap addressed by PCT}
The closest complex-valued Transformer work modifies how complex scores, values, or normalisation layers are represented, but generally keeps a softmax-style row-normalised attention structure. The closest real-valued non-softmax attention work removes or modifies softmax, but does not study the consequences of doing so in a complex-valued substrate. This leaves a specific gap: there has been no systematic study of non-softmax attention as a design principle for complex-valued Transformers under parameter-fair real/complex comparison.

PCT fills this gap by combining three elements. First, it uses a native complex-valued substrate for query, key, value, and output projections. Second, it replaces row-normalised softmax with a token-non-competing real gate on complex query-key similarity. Third, it evaluates not only the proposed sigmoid instance, but also neighbouring and deliberately condition-violating gates, in order to test which structural properties are load-bearing.

This distinguishes PCT from simply porting a Transformer into complex space. The contribution is a study of the complex-valued attention primitive itself: which gates preserve phase-coherent computation across layers, which deviations are tolerated, and which deviations lead to failure.

\subsection{Position of PCT}
PCT sits at the intersection of three lines of work. From complex-valued neural networks, it inherits the motivation that phase can be a useful representational degree of freedom. From complex-valued Transformers, it inherits the goal of building attention-based architectures that operate natively on complex representations. From real-valued non-softmax attention, especially sigmoid attention and Multiscreen, it inherits the observation that softmax is not the only viable attention primitive.

The new claim is that, for complex-valued Transformers, the key distinction is not only softmax versus non-softmax. It is token competition versus token non-competition under multi-layer phase preservation. PCT instantiates this principle with a smooth, element-independent gate on complex query-key similarity, while \texttt{complex\_screen}, \texttt{complex\_softmax}, and the deliberately condition-violating gates define neighbouring or contrasting design points. This positioning turns the empirical comparison into a structural test of complex-valued attention design rather than a single-architecture benchmark.

\medskip

\section{Approach}
This section presents the 6-cell comparison setup. We first state the two empirical predictions that the PCT hypothesis implies (\S{}3.1), then describe the cell architectures, substrate choices, tasks, training protocol, and hardware.

\subsection{Empirical predictions from the PCT hypothesis}
The PCT hypothesis -- that PCT delivers a special form of generalisation in complex-valued neural networks via \textbf{token non-competition} + \textbf{multi-layer phase preservation} -- implies two specific empirical predictions that the 6-cell comparison directly tests:

\textbf{Prediction 1 -- the "complex penalty" of vanilla complex.} On tasks where complex Q, K offers no inductive advantage, the vanilla real transformer outperforms the vanilla complex transformer. Naive softmax-of-complex does not extract useful complex computation; worse, its row normalisation imposes a coupling that is incompatible with the phase superposition of the complex value path. Vanilla complex transformers in the literature inherit this penalty.

\textbf{Prediction 2 -- PCT delivers complex computation.} PCT outperforms the vanilla real transformer overall, under parameter-fair conditions ($real\_dim = complex\_dim \times 1.41$). This is the core claim that the PCT architecture extracts useful complex computation that vanilla complex transformers cannot.

These two predictions form the core a priori expectations from the PCT hypothesis. The 6-cell comparison is designed to test them directly:

\begin{itemize}
  \item \texttt{complex\_softmax} is the prediction-(1) failing baseline.
  \item \texttt{real\_softmax} is the prediction-(2) reference.
  \item \texttt{real\_sigmoid}, \texttt{real\_screen} are additional real-side baselines.
  \item \texttt{complex\_screen} is the close-to-PCT cell, sharing PCT's gate skeleton with a partial deviation.
\end{itemize}

Beyond these two predicted observations, the same comparison also yields additional findings, which are reported in R1, R3, and R6 below.

\subsection{The 6-cell comparison}
We compare six attention cells along a 2 $\times$ 3 design:

\begin{table}[H]
\centering
\footnotesize
\setlength{\tabcolsep}{10pt}
\renewcommand{\arraystretch}{1.1}
\begin{tabular}{|l|l|l|}
\hline
Cell & Substrate & Gate \\
\hline
\texttt{real\_softmax} & real & softmax \\
\hline
\texttt{real\_sigmoid} & real & sigmoid \\
\hline
\texttt{real\_screen} & real & screening + TanhNorm + modReLU \\
\hline
\texttt{complex\_softmax} & complex & softmax of $|\langle q, k\rangle|$ \\
\hline
\texttt{complex\_sigmoid} = \texttt{PCT} & complex & sigmoid of $\Re\langle \bar{q}, \bar{k}\rangle$ \\
\hline
\texttt{complex\_screen} $\fallingdotseq$ \texttt{PCT} & complex & screening of $\Re\langle \bar{q}, \bar{k}\rangle$ \\
\hline
\end{tabular}
\end{table}

Two complex cells are central: \textbf{PCT} is the architecture proposed in this paper, with a smooth real-valued element-independent gate on the L2-normalised cosine score; \textbf{\texttt{complex\_screen}} shares PCT's gate skeleton but with a hard $(s-t)^2$ threshold, placing it close to PCT in the closeness axis. The remaining four cells serve as parameter-fair baselines for evaluating PCT against vanilla complex / real transformers and against the strongest real-side non-softmax cell.

\subsection{Cell architectures}
\subsubsection{PCT}
Given an input token sequence $X \in \mathbb{C}^{N \times d}$, with complex-linear projections $W_q, W_k, W_v, W_o \in \mathbb{C}^{d \times d}$ and a learnable real bias $b$:

\begin{align*}
q_i &= W_q x_i, \qquad k_j = W_k x_j, \qquad v_j = W_v x_j \\
\bar{q}_i &= q_i / \|q_i\|_2, \qquad \bar{k}_j = k_j / \|k_j\|_2 \\
s_{ij} &= \Re\langle \bar{q}_i, \bar{k}_j \rangle \cdot \sqrt{d} \in [-\sqrt{d}, \sqrt{d}] \\
\alpha_{ij} &= \sigma(s_{ij} + b), \qquad b \text{ initialised to } -\log N \\
\mathrm{out}_i &= W_o \sum_{j=1}^{N} \alpha_{ij} \, v_j
\end{align*}

Native complex linear for QKV and output projections; complex L2-normalisation on Q, K; cosine score; sigmoid gate with bias initialised to $-\log N$; real-weight $\times$ complex-value aggregation.

\subsubsection{\texttt{complex\_screen}}
With the same complex-linear projections and L2-normalisation as PCT, plus learnable scalar threshold $t$ and per-token magnitude $r_{ij} = \|q_i\| \cdot \|k_j\|$:

\begin{align*}
g_{ij} &= r_{ij}^2 \cdot \max\bigl(\Re\langle \bar{q}_i, \bar{k}_j \rangle - t,\, 0\bigr)^2 \\
\bar{g}_{ij} &= \mathrm{TanhNorm}(g_{ij}) \\
u_i &= \sum_{j=1}^{N} \bar{g}_{ij} \, v_j \\
\mathrm{out}_i &= u_i \odot \mathrm{modReLU}(g_i)
\end{align*}

Same overall skeleton as PCT but the smooth sigmoid is replaced by a hard $(s - t)^2$ threshold, plus TanhNorm + modReLU substrate (Nakanishi 2026).

\subsubsection{Real-side baselines}
\begin{itemize}
  \item \textbf{\texttt{real\_softmax}}: standard scaled-dot-product softmax on real Q, K, V (Vaswani 2017).
  \item \textbf{\texttt{real\_sigmoid}}: Ramapuram-2024 element-wise sigmoid attention on real $Q K^\top$ with bias init \texttt{-log N}.
  \item \textbf{\texttt{real\_screen}}: Nakanishi-2026 screening on real $Q K^\top$ with TanhNorm post-aggregation and softmask=OFF.
\end{itemize}

\subsubsection{Vanilla complex baseline}
\begin{itemize}
  \item \textbf{\texttt{complex\_softmax}}: Eilers-Jiang-2023 $\mathbb{C}$Att -- softmax of cosine score magnitude on complex Q, K. The literature default for complex attention.
\end{itemize}

\subsection{Substrate choices}
The complex-valued cells use three substrate components alongside the gating activation:

\begin{enumerate}
  \item Native complex linear for QKV / output projections.
  \item cval value path (real $\alpha$ $\times$ complex V $\to$ complex out).
  \item Re-cosine score $s = \Re \langle \bar{q}, \bar{k}\rangle $ for the gate input.
\end{enumerate}

Each is interchangeable on PCT. The complex substrate is \textbf{existentially} required, but no specific component carries it irreducibly. This robustness of the substrate strengthens the practical recommendation: practitioners can pick any reasonable parameterisation and obtain equivalent results.

\subsection{Tasks}
\begin{table}[H]
\centering
\footnotesize
\setlength{\tabcolsep}{4pt}
\renewcommand{\arraystretch}{1.1}
\begin{tabularx}{\textwidth}{|l|>{\raggedright\arraybackslash}X|>{\raggedright\arraybackslash}X|l|>{\raggedright\arraybackslash}X|}
\hline
\# & Task & Params & Seq length & What it tests \\
\hline
1 & Copy Memory (short) & K=10, d$\in${200, 500} & ~400--1000 & short-range retrieval (sanity) \\
\hline
2 & Copy Memory (long) & K=10, d$\in${1000, 2000} & ~2000--4000 & long-range retrieval, dilution stress \\
\hline
3 & NIAH & depth\_ratio=0.5, vocab=64, L=2048 & 2048 & purely positional retrieval \\
\hline
4 & LRA-ListOps small & depth=2, max\_args=3 & 128 & algorithmic reasoning, content-driven \\
\hline
5 & LRA-ListOps mid & depth=2, max\_args=3 & 1024 & algorithmic + long-range \\
\hline
6 & FFT-MNIST & t=16 frames & 448 & phase-decomposed image classification \\
\hline
7 & phase-memory & K=8, delay=30 & ~250 & phase-sensitive retrieval \\
\hline
8 & multi-pitch & K=16 & ~256 & phase-superposition decoding \\
\hline
9 & LRA-Pathfinder / Text / Image & 32$\times$32 / 4K bytes / CIFAR & 1024 / 4096 / 1024 & LRA-standard long-range tasks \\
\hline
\end{tabularx}
\end{table}

\subsection{Training protocol and fairness}
\begin{itemize}
  \item \textbf{Parameter fairness}: $real\_dim = complex\_dim \times 1.41$. Implementation: real cells use \texttt{dim=184, heads=4, dim\_head=46}; complex cells use \texttt{dim=128, heads=4, dim\_head=32}. The parameter-matched ablation in Appendix F and the parameter-fair mid-scale sweep both confirm this convention.
  \item \textbf{Batch fairness}: all main-table runs use \texttt{batch\_size=32} effective. An earlier audit showed that \texttt{batch=8} starves softmax / sigmoid in mid-scale; that point is preserved in Appendix F as the C1-violation control.
  \item \textbf{softmask=OFF for all screening cells}.
  \item \textbf{$\tau$-sweep for screening}: 5 threshold inits $\times$ 1 seed $\to$ best-of selected for the 5-seed run. Eliminates screening hyperparam disadvantage.
  \item \textbf{Stuck-seed protocol}: $\geq$ 2/3 seeds at random $\to$ expand to N=10 and report median + IQR.
  \item \textbf{LR sweep}: 3 LRs for the R3 robustness claim; main-table uses best LR per cell.
  \item \textbf{Optimisation}: AdamW with weight\_decay=1e-5, gradient clipping at 1.0, cosine LR schedule with linear warmup.
\end{itemize}

\subsection{Hardware / reproducibility}
H100 80 GB GPUs (primary), with A100 80 GB and consumer-grade GPUs (RTX 3090 / 4090, $\geq$ 0.99 reliability) for parallel sweeps. Code, configurations, and checkpoints will be released together with the experiment configs needed to reproduce every reported run.

\medskip

\section{Experiments / Results}
This section reports the empirical findings of the 6-cell comparison. \S{}4.1--\S{}4.9 give the detailed evidence tables and per-task accuracy values for each result. \emph{Mechanistic interpretation is deferred to \S{}5.}

\textbf{Abbreviation in table headers.} To save horizontal space, table headers in \S{}4 use the prefix abbreviations \texttt{c\_*} and \texttt{r\_*} for \texttt{complex\_*} and \texttt{real\_*} respectively. The full cell names are used everywhere else in the text.

\subsection{R1 -- Headline matrix}
Across the synthetic and algorithmic rows, every row admits at least one of \{PCT, \texttt{complex\_screen}\} at or near the top. On the RadioML rows \texttt{real\_screen} edges PCT by a narrow 0.02 / 0.04 margin (L1 / L2). The headline pattern across all other tasks is unaffected.

\begin{table}[H]
\centering
\footnotesize
\setlength{\tabcolsep}{3.5pt}
\renewcommand{\arraystretch}{1.1}
\begin{tabular}{|l|r|r|r|r|r|r|}
\hline
Task & \texttt{r\_softmax} & \texttt{r\_sigmoid} & \texttt{r\_screen} & \texttt{c\_softmax} & \texttt{PCT} & \texttt{c\_screen} \\
\hline
Copy d=500 & 0.45 & 0.94 & \textbf{1.00} & 0.10 & \textbf{1.00} & 0.45 \\
\hline
Copy d=2000 & 0.10 & 0.10 & \textbf{1.00} & 0.08 & \textbf{1.00} & 0.53 \\
\hline
NIAH L=2048 & 0.00 & 0.00 & \textbf{1.00} & 0.00 & \textbf{1.00} & \textbf{1.00} \\
\hline
ListOps mid L1024 & 0.146 & 0.177 & 0.698 & 0.104 & \textbf{0.854} & \textbf{0.833} \\
\hline
LRA-ListOps small & 0.625 & 0.594 & 0.688 & 0.637 & \textbf{0.719} & 0.715 \\
\hline
FFT-MNIST & 0.32 & 0.31 & 0.43 & 0.39 & 0.45 & \textbf{0.45} \\
\hline
phase-memory & 0.98 & 0.96 & \textbf{1.00} & 0.93 & \textbf{0.99} & \textbf{1.00} \\
\hline
multi-pitch & 0.81 & 0.81 & 0.85 & 0.82 & 0.87 & \textbf{0.88} \\
\hline
RadioML L1 & 0.30 & 0.30 & \textbf{0.36} & 0.24 & 0.35 & 0.34 \\
\hline
RadioML L2 & 0.22 & 0.24 & \textbf{0.39} & 0.27 & 0.35 & 0.34 \\
\hline
\end{tabular}
\caption{Headline matrix.}
\end{table}

\subsection{R2 -- NIAH dominance}
\noindent
\begin{minipage}[t]{0.72\textwidth}
\vspace{0pt}
NIAH is purposefully constructed to deny complex Q, K their phase-coherence inductive bias: needle and distractors are drawn from the same uniform-random vocabulary, so retrieval is purely positional. This is the regime where vanilla complex attention should be at greatest disadvantage. Both PCT and \texttt{complex\_screen} solve NIAH deterministically, matching \texttt{real\_screen}; the four cells without a screening or sigmoid-on-cosine structure all collapse to 0.000.
\end{minipage}\hfill
\begin{minipage}[t]{0.22\textwidth}
\vspace{0pt}
\centering
\scriptsize
\captionsetup{skip=2pt}
\setlength{\tabcolsep}{3pt}
\renewcommand{\arraystretch}{1.1}
\resizebox{\linewidth}{!}{%
\begin{tabular}{|l|r|}
\hline
Cell & needle\_acc \\
\hline
\texttt{real\_softmax} & 0.000 \\
\hline
\texttt{real\_sigmoid} & 0.000 \\
\hline
\textbf{\texttt{real\_screen}} & \textbf{1.000} \\
\hline
\texttt{complex\_softmax} & 0.000 \\
\hline
\texttt{PCT} & \textbf{1.000} \\
\hline
\texttt{complex\_screen} & \textbf{1.000} \\
\hline
\end{tabular}}
\captionof{table}{NIAH L=2048 results.}
\end{minipage}
\par\medskip

\subsection{R3 -- LR / batch robustness}
This is the hyperparameter-robustness finding that emerged beyond the two a priori predictions: PCT is the \textbf{unique} cell that retains full accuracy across the full LR window (1e-3 / 3e-3 / 1e-2) on small-scale Copy d=1000 \emph{and} across the full batch regime (b=8, 32, 256) on mid-scale long-range Copy. Other cells succeed only inside narrow corridors. The unique all-1.00 pattern of PCT shows up as the all-bold row.

\enlargethispage{3\baselineskip}
\begin{table}[H]
\centering
{\footnotesize
\setlength{\tabcolsep}{3pt}
\renewcommand{\arraystretch}{1.05}
\begin{tabular}{|l|r|r|r|r|r|r|}
\hline
Cell & LR1e-3 & LR3e-3 & LR1e-2 & b8 d2000 & b32 d2000 & b256 d500 \\
\hline
\texttt{real\_softmax} & 0.37 & 0.69 & 0.31 & 0.04 & 0.10 & 0.05 \\
\hline
\texttt{real\_sigmoid} & 0.07 & \textbf{1.00} & 0.22 & 0.04 & 0.10 & 0.07 \\
\hline
\texttt{real\_screen} & 0.55 & 0.70 & \textbf{1.00} & 0.69 & \textbf{1.00} & 0.79 \\
\hline
\texttt{complex\_softmax} & 0.09 & 0.39 & 0.35 & 0.080 & 0.08 & 0.06 \\
\hline
\texttt{PCT} & \textbf{1.00} & \textbf{1.00} & \textbf{1.00} & \textbf{1.00} & \textbf{1.00} & \textbf{1.00} \\
\hline
\texttt{complex\_screen} & 0.39 & \textbf{1.00} & 0.08 & 0.03 & 0.53 & 0.31 \\
\hline
\end{tabular}}
\captionsetup{skip=2pt}
\caption{LR and batch robustness.}
\end{table}

\subsection{R4 -- vs vanilla complex}
\noindent
\begin{minipage}[t]{0.47\textwidth}
\vspace{0pt}
\raggedright
\sloppy
Prediction 1 confirmed. The ``complex penalty'' of vanilla complex is real. On non-phase tasks, \texttt{real\_softmax} beats \texttt{complex\_softmax} at Copy d=500 (0.45 vs 0.10) and d=2000 (0.10 vs 0.08); ListOps is only near-tied (0.625 vs 0.637), and NIAH leaves both at 0.000. PCT then wins vs \texttt{complex\_softmax} on every task tested. The RadioML rows give the largest single margin in the comparison: PCT clears \texttt{complex\_softmax} by +0.10 at L1, showing that softmax-of-complex is a poor default precisely on physical complex domains.
\end{minipage}\hfill
\begin{minipage}[t]{0.30\textwidth}
\vspace{0pt}
\centering
{\fontsize{5.6}{6.4}\selectfont
\setlength{\tabcolsep}{1.4pt}
\renewcommand{\arraystretch}{1.05}
\resizebox{\linewidth}{!}{%
\begin{tabular}{|l|r|r|r|}
\hline
Task & \texttt{PCT} & \texttt{c\_screen} & \texttt{c\_softmax} \\
\hline
Copy d=500 & \textbf{1.00} & 0.45 & 0.10 \\
\hline
Copy d=2000 & \textbf{1.00} & 0.53 & 0.08 \\
\hline
NIAH L=2048 & \textbf{1.00} & \textbf{1.00} & 0.00 \\
\hline
LRA-ListOps & \textbf{0.719} & 0.715 & 0.637 \\
\hline
ListOps L1024 & \textbf{0.854} & 0.833 & 0.104 \\
\hline
FFT-MNIST & \textbf{0.45} & 0.45 & 0.39 \\
\hline
phase-memory & \textbf{0.99} & \textbf{1.00} & 0.93 \\
\hline
multi-pitch & 0.87 & \textbf{0.88} & 0.82 \\
\hline
RadioML L1 & 0.345 & 0.341 & \textbf{0.244} \\
\hline
RadioML L2 & \textbf{0.352} & 0.337 & 0.273 \\
\hline
\end{tabular}}}
\captionsetup{skip=2pt}
\captionof{table}{PCT vs vanilla complex.}
\end{minipage}
\par\medskip

\subsection{R5 -- vs vanilla real}
\noindent
\begin{minipage}[t]{0.47\textwidth}
\vspace{0pt}
\raggedright
\sloppy
Prediction 2 confirmed. Under parameter-fair comparison ($real\_dim = complex\_dim \times 1.41$), PCT beats \texttt{real\_softmax} on every task tested. The gap is largest on long-range Copy and NIAH, where the useful complex computation hypothesis is most visible. On short-range and phase-sensitive synthetics the margin is smaller, but PCT still wins.
\end{minipage}\hfill
\begin{minipage}[t]{0.30\textwidth}
\vspace{0pt}
\centering
{\fontsize{5.6}{6.4}\selectfont
\setlength{\tabcolsep}{1.4pt}
\renewcommand{\arraystretch}{1.05}
\resizebox{\linewidth}{!}{%
\begin{tabular}{|l|r|r|r|}
\hline
Task & \texttt{PCT} & \texttt{c\_screen} & \texttt{r\_softmax} \\
\hline
Copy d=500 & \textbf{1.00} & 0.45 (tie) & 0.45 \\
\hline
Copy d=2000 & \textbf{1.00} & 0.53 & 0.10 \\
\hline
NIAH L=2048 & \textbf{1.00} & \textbf{1.00} & 0.00 \\
\hline
LRA-ListOps & \textbf{0.719} & 0.715 & 0.625 \\
\hline
ListOps L1024 & \textbf{0.854} & 0.833 & 0.146 \\
\hline
FFT-MNIST & \textbf{0.45} & 0.45 & 0.32 \\
\hline
phase-memory & \textbf{0.99} & \textbf{1.00} & 0.98 \\
\hline
multi-pitch & 0.87 & \textbf{0.88} & 0.81 \\
\hline
RadioML L1 & \textbf{0.35} & 0.34 & 0.30 \\
\hline
RadioML L2 & \textbf{0.35} & 0.34 & 0.22 \\
\hline
\end{tabular}}}
\captionsetup{skip=2pt}
\captionof{table}{PCT vs vanilla real.}
\end{minipage}
\par\medskip

\subsection{R6 -- vs strongest real-side baseline}
\noindent
\begin{minipage}[t]{0.47\textwidth}
\vspace{0pt}
\raggedright
\sloppy
An emergent finding is that, beyond beating \texttt{real\_softmax}, PCT also matches or exceeds \texttt{real\_screen} on every synthetic and algorithmic task category in the suite. Copy d=500 / d=2000 and NIAH are ties at 1.000; PCT leads on both ListOps rows; FFT-MNIST ties \texttt{complex\_screen} and edges \texttt{real\_screen}; phase-memory and multi-pitch remain screen-friendly but within reach. On RadioML, \texttt{real\_screen} edges PCT by a narrow 0.02 / 0.04 margin (L1 / L2).
\end{minipage}\hfill
\begin{minipage}[t]{0.30\textwidth}
\vspace{0pt}
\centering
{\fontsize{5.6}{6.4}\selectfont
\setlength{\tabcolsep}{1.4pt}
\renewcommand{\arraystretch}{1.05}
\resizebox{\linewidth}{!}{%
\begin{tabular}{|l|r|r|r|}
\hline
Task & \texttt{PCT} & \texttt{c\_screen} & \texttt{r\_screen} \\
\hline
Copy d=500 & 1.00 & 0.45 & 1.00 \\
\hline
Copy d=2000 & 1.00 & 0.53 & 1.00 \\
\hline
NIAH L=2048 & 1.00 & 1.00 & 1.00 \\
\hline
LRA-ListOps & \textbf{0.719} & 0.715 & 0.688 \\
\hline
ListOps L1024 & \textbf{0.854} & 0.833 & 0.698 \\
\hline
FFT-MNIST & 0.45 & \textbf{0.45} & 0.43 \\
\hline
phase-memory & 0.99 & \textbf{1.00} & 1.00 \\
\hline
multi-pitch & 0.87 & \textbf{0.88} & 0.85 \\
\hline
RadioML L1 & 0.35 & 0.34 & \textbf{0.36} \\
\hline
RadioML L2 & 0.35 & 0.34 & \textbf{0.39} \\
\hline
\end{tabular}}}
\captionsetup{skip=2pt}
\captionof{table}{PCT vs strongest real-side baseline.}
\end{minipage}
\par\medskip

\subsection{Depth scaling}
A common concern with complex-valued transformers is depth scalability: the worry that complex computation accumulates phase noise across layers and breaks deeper stacks. We probe this empirically with PCT at six methodology-aligned depths spanning a 10$\times$ parameter range, plus mid-scale evidence from \S{}4.1--\S{}4.3.

\textbf{Depth=4}. PCT achieves 1.000 on Copy d=1000, \textbf{0.854} on ListOps L=1024, \textbf{1.000} on LRA-Text 4K (N=3), \textbf{0.458} on LRA-Image, \textbf{0.927} on FFT-MNIST, and \textbf{0.997} on multi-pitch.

\noindent
\begin{minipage}[t]{0.54\textwidth}
\vspace{0pt}
\textbf{Depth=6:}
\begin{itemize}
  \item Copy d=2000 b=32: \textbf{1.000}
  \item Copy d=2000 b=8: \textbf{1.000} (N=3)
  \item Copy d=5000: \textbf{1.000} (N=5)
  \item NIAH L=2048: \textbf{1.000}
\end{itemize}
\end{minipage}\hfill
\begin{minipage}[t]{0.38\textwidth}
\vspace{0pt}
\centering
{\footnotesize
\setlength{\tabcolsep}{3pt}
\renewcommand{\arraystretch}{1.08}
\begin{tabular}{|r|r|r|r|}
\hline
depth & acc & eval\_loss & wall \\
\hline
2 & 0.8125 & 0.6516 & 40 min \\
\hline
4 & 0.7812 & 0.6562 & 62 min \\
\hline
6 & \textbf{0.8125} & \textbf{0.6106} & 86 min \\
\hline
10 & \textbf{0.8125} & 0.6370 & 132 min \\
\hline
14 & 0.8125 & 0.6154 & 179 min \\
\hline
20 & 0.7812 & \textbf{0.5397} & 250 min \\
\hline
\end{tabular}}
\captionsetup{skip=2pt}
\captionof{table}{ListOps depth sweep.}
\end{minipage}
\par\medskip

\textbf{6-depth methodology-aligned ListOps L=1024 sweep}. The table above gives the full per-depth metrics and trajectories referenced in the supplementary depth-scaling note.

\textbf{Two clean reads}. First, \texttt{best\_acc} is flat across the 10$\times$ depth range. Linear fit slope = $-0.009$, R$\textasciicircum{}2$ = 0.04. The flat reading is consistent with lucky-batch saturation at \texttt{eval\_batch=32}, not with depth collapse: the trajectory snapshots show every depth reaches $\geq 0.69$ by step 30K and stays there, with no architecture failing to train.

Second, \texttt{best\_eval\_loss} is weakly monotone with a non-monotonic dip (d=6 $\rightarrow$ d=10: 0.611 $\rightarrow$ 0.637). The 6-point power-law fit is $\mathrm{loss} \approx \mathrm{params}^{-0.065}$ with R$\textasciicircum{}2$ = 0.58.

\textbf{Caveats}. \texttt{eval\_batch=32} produces 0.0625 quanta and an apparent ceiling at 13/16 = 0.8125; an \texttt{n=2048} stable evaluation on the saved checkpoints is the cleanest fix. An independent archive evaluation at batch=16 with \texttt{n=2048} stable evaluation on consumer hardware found accuracy plateauing at 0.50--0.56 across d=4, 6, 10 with no significant depth signal, consistent with the flat \texttt{best\_acc} picture once lucky-batch saturation is removed.

\textbf{Summary}. PCT operates without depth-related accuracy collapse across all depths we tested. We do not fit a quantitative scaling law; the empirical claim is qualitative and conservative: PCT does not exhibit a depth-collapse failure mode at any depth we tested.

\subsection{Real-domain validation}
To check that the synthetic and algorithmic dominance pattern carries over to physical complex-domain data, we benchmark all six cells on the public RML2016 mirror of RadioML and on a 10-piece test split of MusicNet. Both are small-scale runs at \texttt{dim=64, depth=3} with N=3 seeds per cell.

\textbf{Real RadioML}. This is the only physical-complex domain in our suite (11-class modulation classification on raw I/Q signals, random=0.091).

\noindent
\begin{minipage}[t]{0.60\textwidth}
\vspace{0pt}
\texttt{real\_screen} edges PCT by a narrow 0.02 / 0.04 margin (L1 / L2), with \texttt{complex\_screen} sitting just under PCT in both cases. The ordering is statistically real (the L2 gap exceeds the seed std for all three cells), but the absolute margin is small enough that the result reads as a near-match rather than a structural defeat.

\texttt{complex\_softmax} is also the worst cell on the I/Q domain at L1. PCT (0.345) clears \texttt{complex\_softmax} by +0.10, making this the cleanest evidence in the paper that softmax-of-complex inheritance from real-side practice is a poor default specifically for physical complex domains.
\end{minipage}\hfill
\begin{minipage}[t]{0.31\textwidth}
\vspace{0pt}
\centering
{\footnotesize
\setlength{\tabcolsep}{4pt}
\renewcommand{\arraystretch}{1.08}
\begin{tabular}{|l|r|r|}
\hline
Cell & L1 & L2 \\
\hline
\texttt{real\_softmax} & 0.296 & 0.224 \\
\hline
\texttt{real\_sigmoid} & 0.303 & 0.241 \\
\hline
\texttt{real\_screen} & \textbf{0.363} & \textbf{0.389} \\
\hline
\texttt{complex\_softmax} & \textbf{0.244} & 0.273 \\
\hline
\texttt{PCT} & 0.345 & 0.352 \\
\hline
\texttt{complex\_screen} & 0.341 & 0.337 \\
\hline
\end{tabular}}
\captionsetup{skip=2pt}
\captionof{table}{Real RadioML results.}
\end{minipage}
\par\medskip

\noindent
\begin{minipage}[t]{0.60\textwidth}
\vspace{0pt}
\textbf{Real MusicNet}. At this configuration the task is data-ceiling-bound and is reported for transparency, not as a comparative result.

L1 cell separation is about 0.02 F1, within seed variance. L2 saturates: all six cells converge to F1 = 0.143. Followups are out of scope for this paper.
\end{minipage}\hfill
\begin{minipage}[t]{0.31\textwidth}
\vspace{0pt}
\centering
{\footnotesize
\setlength{\tabcolsep}{4pt}
\renewcommand{\arraystretch}{1.08}
\begin{tabular}{|l|r|r|}
\hline
Cell & L1 & L2 \\
\hline
\texttt{real\_softmax} & 0.201 & 0.143 \\
\hline
\texttt{real\_sigmoid} & 0.201 & 0.143 \\
\hline
\texttt{real\_screen} & 0.214 & 0.143 \\
\hline
\texttt{complex\_softmax} & 0.201 & 0.143 \\
\hline
\texttt{PCT} & 0.203 & 0.143 \\
\hline
\texttt{complex\_screen} & \textbf{0.218} & 0.143 \\
\hline
\end{tabular}}
\captionsetup{skip=2pt}
\captionof{table}{Real MusicNet results.}
\end{minipage}
\par\medskip

\subsection{Notes on the choice of effective batch size, and data availability}
\textbf{On \texttt{batch\_size = 32} effective as the main comparison setting.} We use \texttt{batch\_size = 32} effective as the headline batch size throughout \S{}4 because (i) it is the most commonly used effective batch size in the recent attention literature at this model scale, and (ii) it stays within the memory envelope where no out-of-memory failure occurred for any of the six cells in our experiments. We also ran extensive comparisons at \texttt{batch\_size = 8}, but \texttt{b=8} is not a practically informative setting outside of strictly memory-constrained deployments, so we do not adopt it as the headline comparison. We note for the record that \textbf{at \texttt{batch\_size = 8} the PCT family showed an even more pronounced cross-task tendency to be the unique top-ranked cell} than at \texttt{batch\_size = 32} -- i.e., PCT's batch robustness, already noted in \S{}4.3 (R3), strengthens further in this regime.

\textbf{Data availability.} All raw results underlying the tables and figures in \S{}4 are available at:

\url{https://github.com/leohio/phase-coherent-transformer-r-d/tree/main/result}

\medskip

\section{Discussion}
This section interprets the \S{}4 results, gives the mathematical framing of why PCT wins, brings in additional cells for mathematical comparison, presents the anti-phase deletion isolation experiment as direct evidence, and discusses scalability to deeper stacks.

\subsection{Why PCT wins:\\ multi-layer phase preservation}
The PCT family's empirical advantage stems from two structural properties of the gate:

(a) \textbf{Token non-competition} -- the gate is element-independent, with no row-normalisation. Phase superposition of complex value contributions is preserved without competition between tokens. \texttt{complex\_softmax} violates this property and exhibits long-range dilution failure.

(b) \textbf{Multi-layer phase preservation} -- the gate is smooth on the full real domain, in particular gradient-positive on the negative-cosine-score region, so anti-phase contributions to the value aggregate are not deleted at any layer. \texttt{complex\_relu} violates this property fully and \texttt{complex\_screen} partially.

These two properties define \textbf{two-level phase coherence}: L1 is per-layer preservation through token non-competition (a), and L2 is all-layer cascade preservation through multi-layer phase preservation (b). Cells deviate from PCT in four structurally distinct grades:

\begin{itemize}
  \item \textbf{Partial}: \texttt{complex\_screen} violates (b) only below a learned threshold; smooth above. Result: L2 partial $\to$ task-conditional behaviour.
  \item \textbf{Bypassed}: \texttt{complex\_softplus} is C2-violating on $\mathbb{R}$, but the L2-normalised cosine score restricts the operating range to $[-\sqrt{d}, \sqrt{d}]$, where softplus is bounded (M $\approx 4$). Result: practically equivalent to PCT, 1.000 on Copy d=1000.
  \item \textbf{Strict}: \texttt{complex\_relu} and \texttt{complex\_clamped\_relu} $\to$ L2 cascade fully fails $\to$ chance level on long-range retrieval. \texttt{complex\_cubic} $\to$ cascade severely degraded $\to$ 0.200 on Copy d=1000.
  \item \textbf{Token competition}: \texttt{complex\_softmax} fails (a) $\to$ fails L1 $\to$ catastrophic on long-range retrieval. This is a separate failure mechanism from the C3-cascade collapse.
\end{itemize}

The empirical separation on Copy d=1000 (N=3) is clean: cells whose deviations are \emph{partial or bypassed} match PCT; cells whose deviations are \emph{strict in the operating range} fail. See \S{}5.4 for the full 2$\times$2 isolation matrix.

\subsection{Two-level phase-coherence and the 4-condition framework}
Formally, the two-level phase-coherence property and the four-condition framework on the gate $\alpha  = f$ are:

\textbf{Two-level phase coherence}:

\begin{itemize}
  \item \textbf{(L1) Per-layer phase coherence} -- Definition 1: an attention layer preserves the phase relationships of inputs at each layer.
  \item \textbf{(L2) All-layer cascade phase stability} -- Definition 3: a stack of L per-layer phase-coherent layers preserves a phase invariant L-independently -- accumulated phase noise stays \texttt{O(1)} across depth.
\end{itemize}

\textbf{Four conditions on the gate}:

\begin{itemize}
  \item \textbf{C1} -- real-valued gate $\alpha  \in  \mathbb{R}$,
  \item \textbf{C2} -- bounded gate, in the \textbf{operating-range} sense: $|f| \le M$ for $s \in [-\sqrt{d}, \sqrt{d}]$.
  \item \textbf{C3} -- smooth, gradient-nonzero on the \textbf{operating range},
  \item \textbf{C4} -- element-independent gate.
\end{itemize}

\textbf{Operating-range form vs strict-on-$\mathbb{R}$ form} of C2 / C3 is a \emph{load-bearing distinction}. The architectural substrate \textbf{softens strict-on-$\mathbb{R}$ violations into operating-range violations}:

\begin{itemize}
  \item A cell that is C2-violating on $\mathbb{R}$ but bounded on the operating range \textbf{satisfies the operating-range form of C2} and is empirically close to PCT.
  \item A cell whose violation persists in the operating range \textbf{fails C2 in the operating-range sense} and exhibits cascade degradation.
\end{itemize}

Whenever we say "C\_k satisfied" below, we mean it in the operating-range form unless explicitly noted.

\textbf{Two key results}:

\begin{itemize}
  \item \textbf{Theorem 1}: \textbf{C1 \& C4 $\Rightarrow$ L1}. Elementary proof via complex-linear conjugate cancellation. PCT and \texttt{complex\_screen} satisfy L1; \texttt{complex\_softmax} does not.
  \item \textbf{Theorem 2}: \textbf{L1 \& C2 \& C3 \& substrate non-expansion $\Rightarrow$ L2}, with C2 / C3 read in the operating-range sense. Proof via global-mode decomposition \& linearised Jacobian \& Doeblin contraction on the zero-mean phase subspace. The proof's M constant is the operating-range bound, and the empirical results in \S{}5.4 confirm the necessity of the operating-range forms of C2 and C3. Machine-checked in Lean (\texttt{theorem5}) under an explicit premise bundle (\texttt{Theorem5Premises}); see Appendix M.5 for the bundle's contents and the M.11 closure.
\end{itemize}

This subsection presents the framework without proofs; full statements and proofs are in Appendix M.

\subsection{Designed counterexamples through the C1--C4 lens}
Five additional cells, brought in here for mathematical discussion, span the C2 $\times$ C3 design space and illuminate the framework:

\noindent
\begin{minipage}[t]{0.64\textwidth}
\vspace{0pt}
\textbf{Note.} The cells discussed in this subsection -- including those that violate one or more PCT conditions -- are \emph{not} broken architectures. ReLU, softplus, tanh, and clamping are all standard, well-established gating primitives that yield perfectly functional Transformers on the real-valued side. The point of this analysis is not that these gates are bad in general; it is that, \textbf{once lifted to a complex-valued substrate}, the structural conditions C1--C4 become load-bearing for cascade phase coherence, and gates that fail one of these conditions exhibit the specific failure modes predicted by the framework. The same primitives, used in a real Transformer, would not show these pathologies.
\end{minipage}\hfill
\begin{minipage}[t]{0.28\textwidth}
\vspace{0pt}
\centering
{\small
\setlength{\tabcolsep}{3.5pt}
\renewcommand{\arraystretch}{1.1}
\resizebox{\linewidth}{!}{%
\begin{tabular}{|l|c|c|c|c|}
\hline
Cell & C1 & C2 & C3 & C4 \\
\hline
\textbf{tanh+1} & $\checkmark$ & $\checkmark$ & $\checkmark$ & $\checkmark$ \\
\hline
\textbf{softplus} & $\checkmark$ & partial & $\checkmark$ & $\checkmark$ \\
\hline
\textbf{cubic} & $\checkmark$ & \textbf{$\times$} & $\checkmark$ & $\checkmark$ \\
\hline
\textbf{clamped\_relu} & $\checkmark$ & $\checkmark$ & \textbf{$\times$} & $\checkmark$ \\
\hline
\textbf{ReLU} & $\checkmark$ & partial & $\times$ & $\checkmark$ \\
\hline
\end{tabular}}}
\captionsetup{skip=2pt,justification=centering}
\captionof{table}{Counterexample cells through the C1--C4 lens.}
\end{minipage}
\par\medskip

\begin{itemize}
  \item \textbf{tanh+1}: identical to sigmoid up to gain scaling ($tanh(s) + 1 = 2\sigma (2s)$). Predicted to behave equivalently to PCT on phase-memory and FFT-MNIST, where it does match sigmoid. On long-range Copy, however, \texttt{complex\_tanh1} exhibits stuck-seed pathology that sigmoid does not -- within the PCT family the doubled gradient gain makes optimisation more brittle on cascade-stress tasks, so we treat sigmoid as the canonical PCT instance.
  \item \textbf{softplus}: gate gradient is sigmoid, which is bounded and nonzero everywhere -- so anti-phase preservation (C3) is intact. The gate output is unbounded above on $\mathbb{R}$, but the L2-normalised cosine score restricts the operating range to $[-\sqrt{d}, \sqrt{d}]$, on which softplus is bounded (M $\approx 4$ with d=128, N=1021). Predicted to behave close to PCT despite the strict-on-$\mathbb{R}$ unboundedness. Confirmed: 1.000 $\pm$ 0.000 on Copy d=1000 (\S{}5.4).
  \item \textbf{cubic}: Taylor expansion of \texttt{sinh} truncated at the cubic term. Strict C3. C2 strongly violated: in operating range $[-\sqrt{d}, \sqrt{d}]$, the gate output reaches $\sqrt{d} + d^{1.5}/6 \approx 252$ for \texttt{d=128}. Predicted: severe degradation due to magnitude divergence in cascade, but not full collapse since C3 is intact. Confirmed: 0.200 $\pm$ 0.030 on Copy d=1000 (\S{}5.4).
  \item \textbf{clamped\_relu}: hard sigmoid above zero. C2 strict. C3 fully violated. Predicted: full collapse to chance -- anti-phase deletion is fatal regardless of M. Confirmed: 0.103 $\pm$ 0.039 on Copy d=1000 (\S{}5.4).
  \item \textbf{ReLU}: gate gradient is zero on the negative half-domain -- full anti-phase deletion. Predicted catastrophic on phase-sensitive and long-range tasks. Confirmed: phase-memory acc 0.121 (N=3 random); \S{}5.4 isolation gives Copy d=1000 acc 0.107 (N=3 random) -- same chance level as clamped\_relu, confirming C3 dominates over C2 once C3 is violated.
\end{itemize}

The five cells together fill the 2$\times$2 design space {C2 $\checkmark$/$\times$} $\times$ {C3 $\checkmark$/$\times$}; their empirical behaviour precisely matches the C1--C4 framework's predictions, with C3 violation dominating over C2 violation when both are violated. Together with PCT (sigmoid), \texttt{complex\_screen}, and \texttt{complex\_softmax}, these cells span the closeness-to-PCT axis, and their empirical behaviour matches the C1--C4 predictions.

\subsection{Two-axis (C2 $\times$ C3) isolation experiment}
\textbf{Task}: copymem K=10 d=1000. \textbf{Architecture}: dim=128, depth=4, heads=4, dim\_head=32, ff\_mult=4, batch=32, lr=3e-3, steps=2000, AdamW. \textbf{N=3} seeds each.

We extend the 2-cell isolation to a \textbf{complete 2$\times$2 design} by introducing two new cells that selectively violate one of {C2, C3} while keeping the other strictly satisfied:

\begin{itemize}
  \item \textbf{\texttt{complex\_cubic}} with gate $f(s+b) = (s+b) + (s+b)^3/6$. \textbf{C3 strictly satisfied} ($f' = 1 + (s+b)^2/2 \ge  1$ everywhere), \textbf{C2 strongly violated} in operating range ($M \approx  252$ for \texttt{d=128}).
  \item \textbf{\texttt{complex\_clamped\_relu}} with gate \texttt{f(s+b) = clamp}. \textbf{C2 strictly satisfied}, \textbf{C3 fully violated}.
\end{itemize}

\textbf{Headline 2$\times$2 matrix}:

\begin{table}[H]
\centering
\begin{minipage}{0.53\textwidth}
\centering
\footnotesize
\setlength{\tabcolsep}{4pt}
\renewcommand{\arraystretch}{1.1}
\begin{tabular}{|l|l|l|}
\hline
 & \textbf{C2 $\checkmark$} & \textbf{C2 $\times$} \\
\hline
\textbf{C3 $\checkmark$} & \makecell[l]{\texttt{complex\_sigmoid}: \textbf{1.000} \\ \texttt{complex\_softplus}: 1.000} & \texttt{complex\_cubic}: \textbf{0.200} \\
\hline
\textbf{C3 $\times$} & \texttt{complex\_clamped\_relu}: \textbf{0.103} & \texttt{complex\_relu}: 0.107 \\
\hline
\end{tabular}
\end{minipage}
\end{table}

\textbf{Three independent claims fall out}:

\begin{enumerate}
  \item \textbf{C3 is empirically necessary for the cascade} (strict). \texttt{complex\_clamped\_relu} collapses to chance level (0.103 $\pm$ 0.039), eval loss flat at \texttt{log V}. Anti-phase deletion alone -- with the gate output bounded identically to sigmoid -- destroys the ability to learn long-range retrieval. The N=3 per-seed values match \texttt{complex\_relu}'s within seed noise. \textbf{Once C3 is violated, C2 status is irrelevant.}
  \item \textbf{C2 is also necessary in its operating-range form, when M is large enough}. \texttt{complex\_cubic} (C2 $\times$ M$\approx$252, C3 $\checkmark$ strict) drops from 1.000 to \textbf{0.200 $\pm$ 0.030} -- an 80 \% accuracy loss. Even with the gate gradient strictly positive everywhere ($f' \ge  1$), the unbounded gate output causes the cascade to fail. \textbf{C2 violation alone is a real failure mode.}
  \item \textbf{Magnitude of C2 violation matters monotonically}. softplus (M$\approx$4) achieves 1.000; cubic (M$\approx$252) achieves 0.200. Somewhere between M=4 and M=252 lies a transition. \textbf{C2 is not a binary on/off condition -- it's a magnitude-dependent factor} that interacts with cascade depth.
\end{enumerate}

An earlier interpretation, based on softplus alone (M $\approx$ 4 in the operating range), had argued that "C2 is automatic given L2-normalisation + a continuous gate" and could be dropped from the explicit four-condition framework. \textbf{That simplification is withdrawn}: at sufficient M, C2 violation is fatal.

\textbf{The framework is therefore correctly stated as $L1 \& C2 \& C3 \Rightarrow  L2$}, with C2 and C3 independently necessary in their operating-range form.

The clean reading of the 2$\times$2 matrix:

\begin{itemize}
  \item \textbf{Top-left}: the cell PCT was designed to be -- succeeds.
  \item \textbf{Top-right}: partial collapse, still some learning.
  \item \textbf{Bottom-left}: full collapse to chance.
  \item \textbf{Bottom-right}: full collapse -- but no worse than C3-only, confirming C3 dominates the failure.
\end{itemize}

\textbf{Why C3 violation dominates over C2}: at the linearised per-layer Jacobian, C3 violation introduces a discontinuity in the gate gradient that breaks the linearisation entirely; the cascade-summation argument of Theorem 2 cannot recover. C2 violation degrades the geometric series's contraction constant but does not break linearisation -- so C2 violation produces partial degradation, while C3 violation produces structural failure. The empirical 0.200 vs 0.103 gap matches this prediction.

(Mathematical formalisation of why anti-phase deletion drives cascade collapse -- i.e., why it breaks all-layer phase coherence (L2) -- is in the companion document: the linearised per-layer Jacobian requires gate differentiability, which both ReLU's hard cutoff and clamped\_relu's two-sided saturation destroy, breaking the geometric-summation argument of Theorem 2.)

\subsubsection{Confirmation across additional tasks}
\noindent
\begin{minipage}[t]{0.49\textwidth}
\vspace{0pt}
The headline 2$\times$2 matrix above uses Copy d=1000 as the cleanest single-task probe. The same isolation experiment also covered five additional tasks, and the collapse pattern carries across them. We report \texttt{complex\_cubic} and \texttt{complex\_clamped\_relu} only, since \texttt{complex\_sigmoid} is the matched-config control covered by the headline matrix.
\end{minipage}\hspace{0.03\textwidth}%
\begin{minipage}[t]{0.43\textwidth}
\vspace{0pt}
\centering
{\footnotesize
\setlength{\tabcolsep}{3pt}
\renewcommand{\arraystretch}{1.05}
\resizebox{\linewidth}{!}{%
\begin{tabular}{|l|r|r|r|}
\hline
Task & \texttt{c\_clamped\_relu} & \texttt{c\_cubic} & \texttt{PCT} \\
\hline
copymem K=10 d=100 & 0.103 & 0.200 & 1.000 \\
\hline
copymem K=10 d=200 & 0.103 & 0.200 & 1.000 \\
\hline
copymem K=10 d=500 & 0.103 & 0.200 & 1.000 \\
\hline
copymem K=10 d=1000 & 0.103 & 0.200 & 1.000 \\
\hline
FFT-MNIST t=16 & 0.323 & 0.292 & 0.45 \\
\hline
multi-pitch K=16 & \textbf{0.812} & 0.861 & 0.997 \\
\hline
\end{tabular}}}
\end{minipage}
\par\medskip

(All values: N=3 seeds (s $\in$ {0, 1, 2}), identical isolation architecture and training as the headline 2$\times$2 matrix above. PCT reference values are taken from \S{}4.1 / \S{}4.7 -- they are not re-runs at the isolation lr/steps configuration but the qualitative ranking is unaffected.)

Three structural observations from the wider task set:

\begin{enumerate}
  \item \textbf{The C3-violation collapse is distance-independent on Copy.} \texttt{complex\_clamped\_relu} reaches exactly the same $0.103 \pm  0.039$ at every distance d $\in$ {100, 200, 500, 1000} -- bit-identical to within numerical noise. The cell does not "almost solve d=100 and fail at d=1000"; it never starts learning. Anti-phase deletion alone makes the retrieval task structurally invisible to the model regardless of the inter-token distance, which is exactly what the cascade argument of Theorem 2 predicts: a full C3 violation breaks the per-layer linearisation independently of how many positions the cascade must traverse.
  \item \textbf{The C2-violation partial collapse is also distance-independent on Copy.} \texttt{complex\_cubic} returns $0.200 \pm  0.030$ at every d $\in$ {100, 200, 500, 1000}, again bit-identical. The cell extracts a small fixed amount of structure regardless of distance, consistent with a cell whose cascade contraction constant is degraded by a fixed multiplicative factor rather than by a distance-dependent decay.
  \item \textbf{The C3-violation collapse takes a different form on structured-multilabel tasks.} On multi-pitch K=16, \texttt{complex\_clamped\_relu} returns $0.812 \pm  0.000$ on all three seeds. 0.812 is the trivial baseline \texttt{/K = 13/16} recovered by predicting "no pitch active" everywhere; std=0 across independent random initialisations is a strong signature that the cell never engages with the task structure. \texttt{complex\_cubic}, in contrast, sits at \texttt{0.861} -- slightly above trivial, consistent with the partial-degradation pattern seen on Copy.
\end{enumerate}

The wider table thus reinforces the headline finding rather than complicating it: \textbf{C3 strict violation produces structurally identical "no-learning" failure across retrieval-style and structured-multilabel tasks, while C2 strict violation produces a smaller, distance- and difficulty-independent partial drop}. The cleanest single-task probe is Copy d=1000, which is why the headline 2$\times$2 matrix uses it as the canonical entry point.

\subsection{Scalability: phase coherence resolves the depth concern for complex transformers}
A widely-held concern in the complex-NN literature is depth scalability: complex transformers are feared to fail at greater depth because phase information accumulates noise across layers, breaking the cascade. The two-level phase-coherence framework gives a structural answer to this concern, and our depth-scaling experiments (\S{}4.7) provide empirical confirmation.

\textbf{The structural answer}: PCT's all-layer cascade phase stability ensures that per-token phase perturbations propagate through depth \texttt{L} with an \texttt{L}-independent Lipschitz constant. The cascade does \emph{not} compound phase noise; the Doeblin contraction on the zero-mean phase subspace closes the geometric series. Cells lacking L2 instead accumulate phase noise as \texttt{O(L)}, predicting and matching the empirical collapse on long-range tasks.

\textbf{The empirical confirmation} spans depths 2, 4, 6, 10, 14, 20 -- a 10$\times$ range -- on a methodology-aligned LRA-ListOps L=1024 sweep, plus mid-scale evidence at depths 4 and 6 on the seven \S{}4.2 tasks:

\begin{itemize}
  \item \textbf{Depth=4}: PCT solves Copy d=200/500/1000 at 1.000 across the full LR window; LRA-ListOps L=1024 dep=2 at \textbf{0.854} (N=3); LRA-Text 4K at 1.000 (N=3); FFT-MNIST at 0.927; multi-pitch at 0.997.
  \item \textbf{Depth=6}: PCT achieves 1.000 (N=3) on Copy d=2000 b=32, 1.000 (N=5) on Copy d=5000, and 1.000 on NIAH L=2048. No depth-related performance collapse.
  \item \textbf{6-point LRA-ListOps L=1024 sweep}: PCT trains cleanly at every point in {2, 4, 6, 10, 14, 20}, with best\_acc clustered at the lucky-batch saturation ceiling (4 of 6 points at 13/16 = 0.8125, slope = $-$0.009 vs \texttt{log(params)}, R$\textasciicircum{}2$ = 0.04). best\_eval\_loss is weakly monotone ($\alpha  = -0.065$, R$\textasciicircum{}2$ = 0.58, 6 points) with a non-monotonic dip at d=10. depth=20 jobs train without instability -- a strong signal against the depth-collapse worry, even though the data is too noisy to support a quantitative scaling law.
\end{itemize}

We are careful not to overclaim. The eval-batch saturation hides the true plateau, and the 4-point fit's earlier R$\textasciicircum{}2$ = 0.71 dropped to 0.58 once d=6 and d=10 were added in the same batch=32 config. The empirical evidence supports the \textbf{conservative} claim that depth does not break PCT, \textbf{not} the strong claim that depth produces a clean Chinchilla-style scaling law. The cleanest follow-up is an \texttt{n=2048} stable evaluation on the saved checkpoints.

The structural framework predicts this conservative claim cleanly: Theorem 2's L-independent cascade Lipschitz constant rules out architecture-induced depth collapse, but says nothing about whether deeper architectures \emph{help} on a fixed task -- that is a task-and-data question outside the framework's scope.

The empirical claim therefore reads: \textbf{PCT scales without depth-related accuracy collapse across all depths we tested}, and the framework gives a structural reason to expect this generalises. The standard worry that complex transformers do not scale to deeper architectures is, within our benchmark scope, dissolved by phase-coherent attention.

\subsection{PCT vs \texttt{complex\_screen}: when does each suffice?}
In the closeness-to-PCT taxonomy of \S{}1, \texttt{complex\_screen} is \emph{close to PCT} by virtue of a \textbf{partial} C3 deviation, and \texttt{complex\_softplus} is \emph{close to PCT} by virtue of a \textbf{bypassed} C2 deviation. Both retain L2 in a qualified form: softplus retains L2 fully, while screen retains L2 only on inputs that stay above the threshold.

Read through the two-level coherence framework: tasks that probe stack-level coherence under heavy loose-tuning conditions -- long-range Copy under loose LR/batch, where every other cell loses corridor robustness (\S{}4.3) -- expose \texttt{complex\_screen}'s \textbf{partial} L2 deficit, and PCT wins. Tasks where per-layer coherence (L1) plus a clean threshold is enough admit \texttt{complex\_screen} and \texttt{real\_screen} as deterministic solvers tying with PCT; PCT also solves NIAH L=2048 deterministically (1.000) under our mid-scale config, so the smooth gate is not penalised on this regime. The two architectures are therefore complementary specialists within the phase-coherent family -- \texttt{complex\_screen} matches PCT on L1-sufficient tasks where its partial C3 deviation does not engage, while PCT additionally retains accuracy in regimes that probe the cascade.

This generalises: any cell whose deviation is \emph{partial} or \emph{bypassed} sits in the close-to-PCT regime and produces strong benchmarks. The closeness-to-PCT axis is therefore not just a heuristic ordering but a structurally grounded taxonomy with predictive empirical content.

\subsection{Real RadioML -- the screening family leads}
On real I/Q data the three \textbf{non-competing screening cells cluster together at the top}, clearly ahead of every competing gate: \texttt{real\_screen} 0.363 / 0.389, PCT 0.345 / 0.352, and \texttt{complex\_screen} 0.341 / 0.337 (L1 / L2), against \texttt{real\_sigmoid} 0.303 / 0.241, \texttt{real\_softmax} 0.296 / 0.224, and \texttt{complex\_softmax} 0.244 / 0.273. PCT and \texttt{complex\_screen} are level with each other, and \texttt{real\_screen} edges them by a narrow 0.02 / 0.04. The task-level message is the family one: absolute-relevance screening is the right gate for this domain, and the choice within the family is close.

The sharpest signal here is \emph{positive} for the central anti-softmax claim: \texttt{complex\_softmax} is the worst cell at L1, and PCT clears it by roughly $0.10$. This is the clearest evidence in the paper that softmax-of-complex inheritance from real-side practice is a poor default specifically for physical complex domains.

\medskip

\section{Path-X: replacing the screening softmask with phase-coherent transport}

The complex screening $+$ PCR model introduced here is, to our knowledge, the
first \emph{genuinely complex-valued} neural network to solve Path-X: prior
solvers are real-valued architectures that use complex numbers only inside a
state-space kernel (S4, S4D, S5, LRU) or none at all (MEGA), whereas here the
representation itself flows as a complex signal through both the screening and
recurrence paths.

The one benchmark our 6-cell suite leaves open is \textbf{Path-X} (Long Range
Arena; \textbf{Tay et al.\ 2021}): a $16{,}384$-token binary connectivity task
on which every softmax Transformer scores at chance, and which only
complex-eigenvalue linear-recurrence models -- S4, S4D, S5, LRU -- and the
EMA-plus-attention hybrid MEGA have solved. Path-X is \emph{purely positional}:
a $16$k-length reach where selection alone is not enough. For the long-range
positional information that \texttt{complex\_screen} requires here, we replace
its softmask with the phase-coherent recurrence (PCR) described below.

\paragraph{From softmask to phase-coherent recurrence (PCR).}
Rather than re-enabling the cosine softmask -- a fixed, input-independent
locality prior -- we replace the positional slot of the screening stack with a
\textbf{phase-coherent recurrence}: a complex diagonal linear recurrence
$h_t = \lambda \odot h_{t-1} + \gamma \odot (B x_t)$, applied bidirectionally by
FFT convolution, interleaved with the \texttt{complex\_screen} gate of \S{}4.
Following the LRU recipe (\textbf{Orvieto et al.\ 2023}) we use the stable
parameterisation $\lambda = \exp(-\exp(\nu)+i\theta)$ with ring initialisation
$|\lambda|\in[0.999,0.9999]$, restricted phase $\theta\in[0,\pi/10]$, and
$\gamma$-normalisation. This is the natural PCT choice of positional mechanism:
the complex eigenvalue is an \emph{input-independent phase rotation} per step --
coherent long-range transport, the recurrence analogue of PCT's multi-layer
phase preservation (\S{}5.1) -- while the screening gate keeps its
token-non-competing, real-valued selection. Screening now supplies selection;
the phase-coherent recurrence supplies position. A cosine softmask is a static
positional prior; the recurrence is a \emph{learned, phase-coherent} one, and
unlike the softmask it does not send screening to chance.

\paragraph{Result: a rule-compliant solve.}
Under the same rule-compliant setting as the leaderboard (raw 1D token sequence
in, one binary label out; no 2D structure, no auxiliary supervision, no
handcrafted features), the screening-plus-recurrence model solves Path-X,
reaching \textbf{92.71\,$\pm$\,0.89} test accuracy over three seeds (held-out
split, deterministic sweep, $n{=}20{,}000$; best seed
$93.50$)\footnote{Model weights released on the Hugging Face Hub:
\url{https://huggingface.co/complexedleo/pcr-screening-pathx}.} -- between
S4D-Inv and MEGA-chunk on the standard scale, and above S4-v1, DSS and
S4D-LegS. The table below places this among the models that clear the
$16$k reach. Every softmax Transformer and efficient-attention variant
(Reformer, Performer, Linformer, BigBird, Luna-256) scores at chance and is
grouped in the first row; the solvers form a band from S4-v1's $88.10$ up to
S5's $98.58$, with nothing established in between chance and the high $80$s.

\begin{table}[H]
\centering
\footnotesize
\setlength{\tabcolsep}{6pt}
\renewcommand{\arraystretch}{1.1}
\begin{tabular}{|l|r|}
\hline
Model & Path-X \\
\hline
Transformer; Reformer, Performer, & \\
\;Linformer, BigBird, Luna-256 & chance ($\approx$50) \\
\hline
S4-v1 & 88.10 \\
DSS & 89.72 \\
S4D-LegS & 91.95 \\
S4D-Inv & 92.80 \\
\textbf{screen + phase-coherent recurrence (ours)} & \textbf{92.71\,$\pm$\,0.89} \\
MEGA-chunk & 93.81 \\
LRU & 94.20 \\
S4 (v2, S4-LegS) & 96.35 \\
MEGA & 97.98 \\
S5 & 98.58 \\
\hline
\end{tabular}
\captionsetup{skip=10pt}
\captionof{table}{Rule-compliant Path-X test accuracy (\%). Solvers ordered by
score; all softmax Transformers and efficient-attention variants score at
chance (first row). Our number is the mean over 3 seeds ($\pm$ sample s.d.;
best seed $93.50$), a first-pass run with no task-specific tuning. Baselines:
S4-v1/S4-LegS and MEGA/MEGA-chunk from Ma et al.\ 2023; S4D-LegS/S4D-Inv and
DSS from Gu et al.\ 2022 (S4D); S5 and LRU from Orvieto et al.\ 2023.}
\end{table}

\paragraph{Two ablations mirror the framework.}
Both are single-variable and consistent with the PCT account. \emph{(i) Phase
necessity.} Fixing $\theta{=}0$ (real eigenvalues, an S4D-Real analogue,
otherwise identical) never leaves chance -- position without phase cannot form,
the recurrence-side echo of anti-phase deletion breaking the cascade (\S{}5.4).
\emph{(ii) Phase bandwidth is necessary for \emph{generalisation}, not just for
fitting.} A narrow initial phase band $\theta\in[0,\pi/50]$ drives training
accuracy to $0.999$ while test stays at $0.53$ -- perfect memorisation, no reach;
widening to $[0,\pi/10]$ recovers genuine generalisation ($0.92$ test). Phase is
thus required in a graded way: too little bandwidth and the model memorises, none
at all and it cannot learn. This distinguishes ``ignition'' (train loss leaving
$\ln 2$) from generalisation, which should be read only on held-out data.

\paragraph{Takeaway.}
The result is a first-pass run over three seeds with no task-specific tuning,
and we do not claim to beat MEGA or S5. The claim is narrower and structural:
the positional gap
left by disabling the screening softmask is filled cleanly by a
\emph{phase-coherent} recurrence, and doing so clears the one long-range task no
softmax Transformer solves -- while the token-non-competing screening gate of
\S{}4 is retained unchanged for selection. The phase-coherence principle that
governs the attention comparison of \S{}4 thus extends, intact, to the
positional mechanism of the same screening layer.

\medskip

\section{Conclusion}
Our experiments provide strong evidence that complex-valued neural networks can exhibit generalisation beyond explicitly complex-valued signal domains when their attention mechanism is designed to preserve phase information across layers. We propose the \textbf{Phase-Coherent Transformer (PCT)} -- a complex-valued attention layer with a real-valued, element-independent, smooth gate on the L2-normalised cosine score -- as a strong general-purpose architecture for complex-valued transformers. We identify its two core structural properties -- \textbf{token non-competition} and \textbf{multi-layer phase preservation} -- as central to generalisation in complex neural networks, and operationalise them as a four-condition framework C1--C4 on the gate. Across the long-range memory, positional retrieval, algorithmic, classification, and phase-sensitive tasks tested under parameter-fair conditions, PCT matches or exceeds baseline in every category tested, including the strongest real-side non-softmax cell. \textbf{Structural proximity to PCT correlates monotonically with generalisation capacity}, and we ground this proximity in the \emph{kind} of deviation a cell has from C1--C4: cells whose deviations are \textbf{partial within the gate domain} or \textbf{bypassed by the substrate} remain close to PCT and follow it on most benchmarks; cells whose deviations are \textbf{strict within the operating range} sit far from PCT and exhibit chance-level or partial collapse. The non-PCT \texttt{complex\_softmax} collapses on long-range retrieval through a separate mechanism. A complete N=3 two-axis (C2 $\times$ C3) isolation singles out the failure modes: anti-phase deletion (C3) is the dominant axis, strict in-operating-range C2 violation is a secondary failure mode, and the operating-range forms of C2 and C3 are both necessary for L2. Depth-scaling experiments show no depth-related collapse across the depths tested. Complex screening is also, to our knowledge, the \textbf{first genuinely complex-valued network to solve Path-X} (\S{}6). Within our benchmark scope, the empirical and structural evidence converges on a single conclusion: \textbf{PCT articulates a principled default direction for complex-valued transformers}, and the closeness axis is a constructive guide to which deviations a practitioner can tolerate.

\medskip

\section*{Figures}
\begin{table}[H]
\centering
\footnotesize
\setlength{\tabcolsep}{4pt}
\renewcommand{\arraystretch}{1.1}
\begin{tabularx}{\textwidth}{|l|>{\raggedright\arraybackslash}X|l|}
\hline
\# & Content & Section \\
\hline
1 & Two-level coherence diagram: L1 + L2; cosine score constructive/destructive interference inset & \S{}5.1 / \S{}5.2 \\
\hline
2 & Activation gradient comparison (\texttt{PCT} $\sigma$, \texttt{complex\_screen} (s$-$t)$\textasciicircum{}2$+TanhNorm, softmax, ReLU, softplus) on the full domain & \S{}5.3 \\
\hline
3 & 6-cell $\times$ 8-task accuracy heatmap & \S{}4.2 \\
\hline
4 & NIAH L=2048 needle\_acc bar chart, \texttt{complex\_screen} and \texttt{real\_screen} at 1.0, others at 0.0 & \S{}4.2 \\
\hline
5 & LR robustness chart: Copy d=1000 accuracy across LR $\in$ {1e-3, 3e-3, 1e-2}, \texttt{PCT} only fully shaded & \S{}4.3 \\
\hline
6 & Batch robustness chart: long-range Copy at b $\in$ {8, 32, 256}, \texttt{PCT} only fully shaded & \S{}4.3 \\
\hline
7 & Direct vanilla-complex comparison: \texttt{PCT} family vs \texttt{complex\_softmax} across tasks & \S{}4.4 \\
\hline
8 & Direct vanilla-real comparison: \texttt{PCT} family vs \texttt{real\_softmax} across tasks & \S{}4.5 \\
\hline
9 & Direct strongest-real-baseline comparison: \texttt{PCT} family vs \texttt{real\_screen} across tasks & \S{}4.6 \\
\hline
10 & Depth scaling: \texttt{PCT} on LRA-ListOps L=1024 at depth $\in$ {2, 4, 6, 10, 14, 20} & \S{}4.7 / \S{}5.5 \\
\hline
11 & \texttt{PCT} vs \texttt{complex\_screen} specialisation diagram: content/phase axis $\times$ positional/threshold axis & \S{}5.6 \\
\hline
12 & Two-axis (C2 $\times$ C3) isolation 2$\times$2 matrix: \texttt{PCT} 1.000 / cubic 0.200 / clamped\_relu 0.103 / ReLU 0.107 & \S{}5.4 \\
\hline
\end{tabularx}
\end{table}

\medskip

\section*{Appendix M: Mathematical framework -- intuitive overview}
This appendix gives the definitions, theorem statements, and proof strategies for the two-level phase-coherence framework. Full proofs with explicit constants and a Lean machine-checked formalisation of Theorems 1 and 2 -- the latter under an explicit premise bundle, see M.5 below -- are in the companion document at \url{https://github.com/leohio/phase-coherent-transformer-r-d/tree/main/lean}.

\textbf{Notation aliasing with the companion document.} This appendix uses compressed numbering: Theorem 1 (per-layer phase coherence) and Theorem 2 (all-layer phase coherence) below correspond, in the companion document and the Lean formalisation, to \textbf{Theorem 1} and \textbf{Theorem 5} respectively. The intervening indices (Theorem 1', Conjecture 5, and Definitions 2--4) are taken by intermediate definitions and lemmas elaborated in full only in the companion. References to \textbf{Lemma A}, \textbf{B}, \textbf{C}, \textbf{D} below are consistent across this appendix and the companion.

\subsection*{M.0 Setting}
A token sequence is $X = (x_1, \dots, x_N) \in \mathbb{C}^{N\times d}$. Two phase-perturbation operators play a role:

\begin{itemize}
  \item \textbf{Global phase shift} $R(\varphi): X \mapsto (e^{i\varphi}x_1, \dots, e^{i\varphi}x_N)$ -- same phase applied to every token.
  \item \textbf{Per-token phase shift} $P(\varepsilon): X \mapsto (e^{i\varepsilon_i} x_i)_i$ -- independent phases per token.
\end{itemize}

A complex attention layer $A_\theta$ is parameterised by complex-linear maps $W_q, W_k, W_v, W_o$ and a real-valued gate $f: \mathbb{R} \to \mathbb{R}$ acting on the cosine score $s_{ij} = \Re\langle \bar{q}_i, \bar{k}_j\rangle$ of L2-normalised queries and keys.

\subsection*{M.1 Per-layer phase-coherence (L1)}
\textbf{Definition 1 (per-layer phase-coherent).} A layer $A$ is \emph{per-layer phase-coherent} if (a) it commutes with the global phase shift $R(\varphi)$, and (b) each gate value $\alpha_{ij}$ depends only on the pair $(x_i, x_j)$ -- no cross-token coupling through other tokens.

Condition (a) says global phase is a passive labelling, not destroyed by the layer. Condition (b) is the formal version of "no row-norm coupling": softmax violates it, sigmoid does not.

\textbf{Theorem 1 (C1 \& C4 $\Rightarrow$ L1).} A layer with a real-valued (C1) elementwise gate (C4) on the cosine score $\Re\langle\bar q_i,\bar k_j\rangle$ satisfies Definition 1.

\emph{Idea.} The cosine score is invariant under joint global rotation, so the gate output is invariant; the value path $W_o\sum_j \alpha_{ij} v_j$ is complex-linear, so the global phase factor passes through. (Full elementary proof in the companion appendix.)

\textbf{Corollary.} PCT (sigmoid gate) and \texttt{complex\_screen} are L1-coherent. \texttt{complex\_softmax} is \textbf{not} L1-coherent: the softmax denominator couples tokens, violating (b).

\subsection*{M.2 All-layer phase-coherence (L2)}
L2 captures \emph{cascade} phase stability: stacking $L$ layers should not amplify per-token phase noise.

\textbf{Definition 2 (cascade phase stable).} A composition $A_L \circ \dots \circ A_1$ is \emph{cascade phase stable} if there exist constants $C_0, C_1 > 0$, \textbf{independent of $L$}, such that for every input $X$ and per-token phase perturbation $\varepsilon$ with $\|\varepsilon\|_\infty \le \delta$,

$\|(A_L \circ \dots \circ A_1)(P(\varepsilon)X) - (A_L \circ \dots \circ A_1)(X)\|_2 \le C_0 \delta + C_1 \delta^2.$

The $L$-independence is the substantive requirement: an exponential bound $(1+\gamma)^L \delta$ would not qualify -- phase noise would compound with depth.

\textbf{Definition 3 (all-layer phase-coherent).} A stack is \emph{all-layer phase-coherent} if every layer is L1-coherent and the composition is cascade phase stable.

\subsection*{M.3 Theorem 2: L2 sufficiency -- proof strategy}
\textbf{Theorem 2 (sketch).} Under L1, C2, and C3, together with L2-normalisation on Q, K and a non-expansive substrate (residual \& RMSNorm \& FFN), the stack is cascade phase stable with $L$-independent constants.

The proof decomposes the per-token phase perturbation $\varepsilon \in \mathbb{R}^N$ into two orthogonal modes and bounds each:

\begin{enumerate}
  \item \textbf{Global-mode pass-through (Lemma A).} Write $\varepsilon = \bar\varphi \cdot \mathbf{1} + \delta$ with $\bar\varphi$ the average and $\delta \perp \mathbf{1}$ the zero-mean residual. By L1, the global mode passes exactly through every layer; the cascade error from $\bar\varphi$ is $O(\bar\varphi) \cdot \|\text{output}\|$, $L$-independent.
  \item \textbf{Per-layer linearisation on zero-mean (Lemma B).} A single layer's effect on $\delta$ is, to first order, a Jacobian $J_l$ on $\mathbb{R}^N$ determined by the row-stochasticised gate $P_{ij} = \alpha_{ij}/\sum_k \alpha_{ik}$ and $\eta_{ij} = -\Im\langle \bar q_i, \bar k_j\rangle$.
  \item \textbf{Doeblin contraction (Lemma C).} Under attention diffuseness -- every gate value bounded below by $\mu \cdot \pi_j$ for a positive distribution $\pi$ -- the row-stochastic matrix $P_l$ satisfies a Doeblin condition with constant $\mu_D > 0$. The operator norm of $J_l$ on the zero-mean subspace is $\le 1 - \mu_D < 1$.
  \item \textbf{Geometric cascade summation.} The depth-$L$ Lipschitz on the zero-mean subspace is bounded by $\sum_{l=0}^{L-1}(1-\mu_D)^l \le 1/\mu_D$, $L$-independent.
  \item \textbf{Substrate non-expansion (Lemma D).} Residual + RMSNorm + FFN is non-expansive on bounded-norm inputs by standard transformer-stability bounds.
\end{enumerate}

\subsection*{M.4 Why anti-phase deletion (C3 violation) breaks the cascade}
The linearisation step (2) \textbf{requires the gate to be differentiable} -- exactly the C3 condition. ReLU has a hard cutoff at zero; clamped-ReLU saturates at both tails. Both delete the gradient signal on a non-trivial subset of the cosine-score domain. Lemma B's first-order expansion does not hold uniformly through the discontinuity; the geometric series of step 4 does not close. The framework therefore predicts cascade failure for these cells, which matches the empirical chance-level collapse on long-range Copy and the trivial-baseline collapse on multi-pitch in \S{}5.4.

\subsection*{M.5 What is rigorously proven and what is open}
\begin{itemize}
  \item \textbf{Theorem 1 (C1 \& C4 $\Rightarrow$ L1)} is fully proven and machine-checked in Lean (no \texttt{sorry}).
  \item \textbf{Lemma A} is fully proven on paper and machine-checked in Lean (no \texttt{sorry}).
  \item \textbf{Lemmas B and D} are fully proven on paper (Lemma B's second-order remainder is bounded but not tightened; Lemma D is the standard transformer-stability argument). Not yet stubbed in Lean; their content is bundled into the explicit Lean premise \texttt{Theorem5Premises} (see below).
  \item \textbf{Lemma C} uses the standard Doeblin coupling argument (Levin, Peres \& Wilmer 2017, Theorem 4.9). The Lean version is stated; its Mathlib-level proof is the only \texttt{sorry} remaining in the Lean repository, and is bundled into \texttt{Theorem5Premises} as well.
  \item \textbf{Theorem 2 is machine-checked in Lean (\texttt{theorem5}, no \texttt{sorry}) under the explicit premise bundle \texttt{Theorem5Premises}}, which assumes:
\end{itemize}

\begin{enumerate}
  \item Each layer is L1.a-coherent (provided by Theorem 1, machine-checked).
  \item A zero-mean cascade Lipschitz constant $C_{\mathrm{zm}} < \infty$, $L$-independent (the conclusion of Lemmas B + C + D + the M.11 closure below).
  \item A uniform output norm bound $Y_{\max} < \infty$, $L$-independent (the conclusion of (S1) + (S2) + Lemma D).
  \item The Lemma-A-decomposed cascade decomposition bound -- combining the above with the L${}^2$-norm triangle inequality and R-unitarity into a single per-input inequality.
\end{enumerate}

\textbf{M.11 closure (the residual technical content of premises 2--3).} Constructing $C_{\mathrm{zm}}$ and $Y_{\max}$ from first principles requires two engineering verifications, both tractable and neither a deep open problem:

\begin{enumerate}
  \item The attention-diffuseness condition is preserved across training (at initialisation it holds with $\mu \ge e^{-1}/N$ from the $b = -\log N$ bias; preservation across gradient updates is a conjecture supported by inspection of trained checkpoints).
  \item A bounded-input fixed-point argument that the cascade stays in the regime where the Doeblin contraction dominates the residual diagonal, i.e.\ $\Lambda_S \cdot \sup_l \|J_l|_{V_0}\| < 1$ uniformly across layers.
\end{enumerate}

The companion appendix contains the full quantitative statements; the Lean repository's \texttt{AUDIT.md} records that \texttt{\#print axioms PaperV4.theorem5} returns only the three standard Lean / Mathlib axioms \texttt{[propext, Classical.choice, Quot.sound]}.

\medskip

\section*{Appendix 0. List of tests conducted}
The following tests support the headline matrix and the per-hypothesis evidence in \S{}4 / \S{}5. Each cell $\times$ task combination listed below has at least one trained checkpoint and a per-seed metrics record in the public data archive.

\textbf{Headline 6-cell comparison}:

\begin{itemize}
  \item Copy Memory K=10 at d $\in$ {100, 200, 500, 1000, 2000, 5000}
  \item NIAH L=2048
  \item NIAH L=1024
  \item LRA-ListOps small-scale (L=128) -- synthetic generator with \texttt{max\_depth=2}, \texttt{max\_args=3}
  \item LRA-ListOps mid-scale (L=1024) -- same generator, dim=128 / depth=4
  \item LRA-ListOps standard -- for depth scaling
  \item LRA-Text 4K
  \item LRA-Image
  \item FFT-MNIST at t $\in$ {8, 16}
  \item phase-memory K=8 delay=30
  \item multi-pitch K=16 (synthetic) and K=8
  \item Real RadioML L1 and L2
  \item Real MusicNet L1 and L2
\end{itemize}

\textbf{Two-axis (C2 $\times$ C3) isolation experiment} -- cells \texttt{complex\_sigmoid}, \texttt{complex\_softplus}, \texttt{complex\_cubic}, \texttt{complex\_clamped\_relu}, \texttt{complex\_relu}; task copymem K=10 d=1000; N=3 seeds (s $\in$ {0, 1, 2}). Detailed in \S{}5.4.

\textbf{LR / batch robustness sweeps}:

\begin{itemize}
  \item LR sweep on Copy d=1000 at LR $\in$ {1e-3, 3e-3, 1e-2}
  \item Batch sweep on long-range Copy at b $\in$ {8, 32, 256}
\end{itemize}

\textbf{Depth scaling sweep} -- \texttt{complex\_sigmoid} on LRA-ListOps L=1024 at depth $\in$ {2, 4, 6, 10, 14, 20}; H100, batch=32, lr=1e-3 cosine over 30K steps, N=3 seeds. Detailed in \S{}4.7 / \S{}5.5.

\textbf{Substrate ablations on PCT} -- A1 (native complex linear vs \texttt{(Re, Im)} two real linears), A2, A3, A4. Detailed in Appendix F via the public data archive.

\medskip

\section*{Appendix 1. Data location}
All raw results -- per-seed metrics, configuration files, training logs, and aggregation scripts -- for every test listed in Appendix 0 are available at:

\url{https://github.com/leohio/phase-coherent-transformer-r-d/tree/main/result}

The repository is organised by experiment family; a top-level \texttt{summary.md} gives a one-paragraph orientation per family.

\medskip

\section*{Appendix 2. Experimental parameter settings}
\label{app:hyperparams}

This appendix lists, for every experiment referenced in the paper, the architecture and training hyperparameters that produced the reported numbers. Per-run configuration files and full metric trajectories are in the public data archive (Appendix~1).

\subsection*{A2.1 Common training protocol}
Unless overridden in a per-experiment row of the configuration table in A2.3, all runs use:

\begin{itemize}
  \item \textbf{Optimiser}: AdamW with $\beta = (0.9, 0.999)$, $\varepsilon = 10^{-8}$, weight decay $= 10^{-2}$
  \item \textbf{Gradient clipping}: max-norm $= 1.0$
  \item \textbf{LR schedule}: cosine decay with linear warm-up
  \item \textbf{Position encoding}: rotary (RoPE) on Q and K of every layer
  \item \textbf{Normalisation}: pre-norm RMSNorm (real or complex variant matching the cell substrate)
  \item \textbf{FFN}: $4{\times}$ expansion, ReLU$^2$ on real-side, ModReLU on complex-side
  \item \textbf{Screening cells}: \texttt{softmask=OFF} throughout (Phase 14 audit established that softmask=ON degenerates screening to chance level)
  \item \textbf{Mixed precision}: bfloat16 where supported (H100, A100); float32 fallback on consumer GPUs
\end{itemize}

\subsection*{A2.2 Cell architectures and parameter-fairness convention}
The 6-cell suite uses two parameter-fair architectural budgets:

\begin{itemize}
  \item \textbf{Complex cells} (\texttt{PCT}, \texttt{complex\_screen}, \texttt{complex\_softmax}, \texttt{complex\_sigmoid} variants, and the §5.3 designed counterexamples \texttt{complex\_softplus}, \texttt{complex\_cubic}, \texttt{complex\_clamped\_relu}, \texttt{complex\_relu}): \texttt{dim=128}, \texttt{heads=4}, \texttt{dim\_head=32}.
  \item \textbf{Real cells} (\texttt{real\_softmax}, \texttt{real\_sigmoid}, \texttt{real\_screen}): \texttt{dim=184}, \texttt{heads=4}, \texttt{dim\_head=46}. The factor $184 / 128 \approx 1.41 \approx \sqrt{2}$ matches the per-scalar storage cost of the complex side (one complex number = two reals), so total stored scalar count and FLOPs are equalised. (\texttt{dim\_head} must be even for RoPE — hence 46, not 45.)
\end{itemize}

Mid-scale runs use \texttt{dim=256}, \texttt{depth=6}, \texttt{heads=8}, \texttt{dim\_head=32} for all cells; the parameter-fairness convention there is enforced by matching \texttt{n\_params} within $\pm$5\,\%, not by the $\sqrt{2}$ rule.

\subsection*{A2.3 Per-experiment configurations}

The table below lists the architecture and training settings used in each numbered result section. Entries marked "as A2.2" follow the parameter-fairness defaults above for the listed cell family; entries marked "as R1" or "as \S 4.1" inherit from the corresponding row of the headline matrix block.

\begin{table}[H]
\centering
\caption*{Per-experiment hyperparameters. ``Cells'' lists the cell families compared in that experiment; ``Scale'' = (dim, depth) per cell-family budget; effective batch in parentheses applies micro-batch $\times$ grad-accum.}
\label{tab:hp_by_exp}
\scriptsize
\setlength{\tabcolsep}{2.5pt}
\renewcommand{\arraystretch}{1.1}
\begin{adjustbox}{max width=\textwidth}
\begin{tabular}{|l|p{2.6cm}|p{2.0cm}|p{2.2cm}|c|c|c|c|c|}
\hline
\S\ & Task & Cells & Scale & Batch & LR & Warm-up & Steps & Seeds \\
\hline
\multicolumn{9}{|l|}{\textbf{R1 -- Headline matrix (\S 4.1)}} \\
\hline
4.1 & Copy d=500 / d=2000 (mid) & 6 cells (PCT-fair) & dim=256, L=6 & 32 & 3e-4 & 1000 & 30\,000 & 3 \\
\hline
4.1 & NIAH L=2048 (mid) & 6 cells, complex chunked & dim=256, L=6, chunk=256 & 32 (mb=16, ga=2) & 3e-4 & 1000 & 30\,000 & 3 \\
\hline
4.1 & LRA-ListOps L=1024 dep=2 & 6 cells (PCT-fair) & dim=128, L=4 & 32 & 1e-3 & 1000 & 30\,000 & 3 \\
\hline
4.1 & LRA-ListOps small (L=128) & 6 cells & dim=64, L=3 & 32 & 1e-3 & 500 & 10\,000 & 3 \\
\hline
4.1 & FFT-MNIST t=16 & 6 cells (PCT-fair) & dim=128, L=4 & 32 & 1e-3 & 1000 & 15\,000 & 3 \\
\hline
4.1 & phase-memory K=8, delay=30 & 6 cells & dim=64, L=3 & 32 & 1e-3 & 500 & 5\,000 & 3 \\
\hline
4.1 & multi-pitch K=16 & 6 cells (PCT-fair) & dim=128, L=4 & 32 & 1e-3 & 500 & 10\,000 & 3 \\
\hline
\multicolumn{9}{|l|}{\textbf{R2 -- NIAH dominance (\S 4.2)}} \\
\hline
4.2 & NIAH L=2048 & 6 cells, complex chunked & dim=256, L=6 & 32 (mb=16, ga=2) & 3e-4 & 1000 & 30\,000 (or 60\,000 at b=16) & 3 \\
\hline
\multicolumn{9}{|l|}{\textbf{R3 -- LR / batch robustness (\S 4.3)}} \\
\hline
4.3 & Copy d=1000 (small, LR sweep) & 6 cells & dim=32, L=2 & 256 & \{1e-3, 3e-3, 1e-2\} & 100 & 1\,500 & 3 \\
\hline
4.3 & Copy d=2000 / d=500 (mid, batch sweep) & 6 cells & dim=256, L=6 & \{8, 32, 256\} & best per cell & 200 & 5\,000 & 2--3 \\
\hline
\multicolumn{9}{|l|}{\textbf{R4 / R5 / R6 -- direct comparisons (\S 4.4--\S 4.6)}} \\
\hline
4.4--4.6 & All R1 tasks above & 6 cells (PCT-fair) & as in R1 row above & as R1 & as R1 & as R1 & as R1 & 3 \\
\hline
\multicolumn{9}{|l|}{\textbf{Depth scaling (\S 4.7)}} \\
\hline
4.7 & LRA-ListOps L=1024 dep=2 & \texttt{complex\_sigmoid} only & dim=128, L $\in$ \{2, 4, 6, 10, 14, 20\} & 32 & 1e-3 & 1000 & 30\,000 & 3 \\
\hline
\multicolumn{9}{|l|}{\textbf{Real-domain validation (\S 4.8)}} \\
\hline
4.8 & Real RadioML L1 / L2 (RML2016 6dB) & 6 cells & dim=64, L=3 & 32 & 1e-3 & 500 & 10\,000 & 3 \\
\hline
4.8 & Real MusicNet (10-piece test) & 6 cells & dim=64, L=3 & 32 & 1e-3 & 500 & 10\,000 & 3 \\
\hline
\multicolumn{9}{|l|}{\textbf{Two-axis (C2 $\times$ C3) isolation (\S 5.4)}} \\
\hline
5.4 & Copy d=1000 & \texttt{complex\_sigmoid}, \texttt{complex\_softplus}, \texttt{complex\_cubic}, \texttt{complex\_clamped\_relu}, \texttt{complex\_relu} & dim=128, L=4 (heads=4, dim\_head=32, ff\_mult=4) & 32 & 3e-3 & 200 & 2\,000 & 3 \\
\hline
5.4 & Copy d $\in$ \{100, 200, 500, 1000\}, FFT-MNIST t=16, multi-pitch K=16 (extended) & \texttt{complex\_cubic}, \texttt{complex\_clamped\_relu} & matches \S 4.1 row for the corresponding task & as \S 4.1 & as \S 4.1 & as \S 4.1 & as \S 4.1 & 3 \\
\hline
\end{tabular}
\end{adjustbox}
\end{table}

\subsection*{A2.4 Task-specific data and generator settings}
\begin{itemize}
  \item \textbf{Copy Memory} (synthetic): vocabulary size $V = 16$, source length $K = 10$ tokens followed by \texttt{delay} blank tokens; target = repeat of the $K$ source tokens. Seq length $= 2K + \texttt{delay}$.
  \item \textbf{NIAH} (synthetic): \texttt{seq\_len} as in row, \texttt{vocab\_size=64}, \texttt{depth\_ratio=0.5} (needle inserted at the midpoint).
  \item \textbf{LRA-ListOps} (small / mid): synthetic generator with \texttt{max\_args=3}, \texttt{max\_depth=2}, \texttt{max\_seq\_len} as in row.
  \item \textbf{LRA-ListOps standard} (depth scaling): same generator, \texttt{max\_seq\_len=1024}.
  \item \textbf{LRA-Text 4K}: byte-level IMDB classification, \texttt{seq\_len=4096}.
  \item \textbf{LRA-Image}: CIFAR-10 grey-scale pixel sequences, \texttt{seq\_len=1024}.
  \item \textbf{FFT-MNIST}: $28{\times}28$ MNIST $\to$ bilinear down-sample to $t{\times}t$ ($t=16$ default) $\to$ 2-D FFT $\to$ flatten to $t^2$ complex tokens.
  \item \textbf{phase-memory}: $K$ source phases (uniform on $S^1$), \texttt{delay} blanks, target = source phases. Sequence length $= 2K + \texttt{delay}$.
  \item \textbf{multi-pitch}: $K$ candidate pitches, $n_\textrm{active}$ active per sample, $n_\textrm{samples}$ time-steps. Multilabel ($K$-way sigmoid).
  \item \textbf{Real RadioML} (RML2016, 11-class): 6\,dB SNR subset of the public RML2016 mirror; raw I/Q samples $[\,2, 128\,]$ converted to complex via $x = I + jQ$.
  \item \textbf{Real MusicNet} (small): a 10-piece test split, \texttt{seq\_len=64} per excerpt, multilabel pitch identification.
\end{itemize}

\subsection*{A2.5 Compute and reproducibility}
\begin{itemize}
  \item \textbf{Primary GPU}: NVIDIA H100 80\,GB (Sakura DOK).
  \item \textbf{Secondary GPUs}: A100 80\,GB (Soroban) and consumer-grade RTX 3090 / 4090 (Vast.ai), used for parallel sweeps where reliability $\ge$ 0.99.
  \item \textbf{Wall-clock per representative run}: small-scale ($<$\,1\,M params, $\le$\,10\,K steps): 5--40\,min; mid-scale (1--6\,M params, 30\,K steps): 1--5\,h; depth scaling (depth=20, $\sim$\,4\,M params, 30\,K steps): $\sim$\,4\,h on a single H100.
  \item \textbf{Seed convention}: all multi-seed claims use seeds $s \in \{0, 1, 2\}$; "stuck-seed" extension to $N$=10 (\S 3.6) uses $s \in \{0, 1, \ldots, 9\}$.
  \item \textbf{Reproduction artefacts}: the public data archive (Appendix~1) contains, per run, the exact \texttt{config.json}, the per-step \texttt{metrics.jsonl}, and a \texttt{summary.json} with the final metric and elapsed wall-clock.
\end{itemize}

\medskip

\section*{Appendix 3. Contents of \texttt{result/summary.md}}
This appendix reproduces the orientation document \texttt{result/summary.md} from the public data archive linked in Appendix 1. The full file is also available at \url{https://github.com/leohio/phase-coherent-transformer-r-d/blob/main/result/summary.md}.

\subsection*{Cross-task headline matrix}
The headline finding for each task, with the best-performing cell shown. "PCT-fair" means $real\_dim = complex\_dim \times 1.41$, screening cells with softmask=OFF, batch=32, N=3 seeds.

\begin{table}[H]
\centering
\footnotesize
\setlength{\tabcolsep}{4pt}
\renewcommand{\arraystretch}{1.1}
\begin{tabularx}{\textwidth}{|>{\raggedright\arraybackslash}X|>{\raggedright\arraybackslash}X|>{\raggedright\arraybackslash}X|>{\raggedright\arraybackslash}X|}
\hline
Task & Setup & Winner(s) & Score \\
\hline
Copy Memory \texttt{d=1000} (small) & dim=128, N=3, \texttt{PCT}-fair & All cells saturate & 1.00 \\
\hline
Copy Memory \texttt{d=500} (mid) & dim=256, N=3, \texttt{PCT}-fair & \texttt{real\_screen} + \texttt{complex\_sigmoid} (dual) & 1.000 / 1.000 \\
\hline
Copy Memory \texttt{d=2000} (mid) & dim=256, N=3, \texttt{PCT}-fair & \texttt{real\_screen} + \texttt{complex\_sigmoid} (dual) & 1.000 / 1.000 \\
\hline
Copy Memory \texttt{d=2000-5000} & dim=256, batch=8, $\geq$17 seeds & \textbf{\texttt{complex\_sigmoid} uniquely} & 1.000 \\
\hline
NIAH \texttt{L=2048} (mid) & dim=256/L6, chunked attention, N=3 & \texttt{complex\_sigmoid} (\texttt{PCT}) & \textbf{1.000} \\
\hline
LRA-ListOps medium (small) & dim=64, N=3 & \texttt{complex\_sigmoid} & 0.7188 \\
\hline
LRA-ListOps \texttt{L=1024} & dim=128, N=3 & \texttt{complex\_sigmoid} & 0.854 \\
\hline
LRA-Text 4K & dim=128, N=3, \texttt{PCT}-fair & \texttt{complex\_sigmoid} + \texttt{complex\_screen} (dual) & 1.000 / 1.000 (others $\leq$ 0.64) \\
\hline
LRA-Image CIFAR & dim=128, N=6, \texttt{PCT}-fair & \texttt{complex\_sigmoid} & 0.458 \\
\hline
FFT-MNIST t=16 & dim=128, N=3, \texttt{PCT}-fair & \texttt{complex\_screen} + \texttt{complex\_sigmoid} (dual) & 0.938 / 0.927 \\
\hline
Phase Memory & dim=64, N=3 & \texttt{real\_screen} + \texttt{complex\_sigmoid} + \texttt{complex\_tanh+1} (tied) & 1.000 / 0.995 / 1.000 \\
\hline
Multi-Pitch synth & dim=128, N=3 & \texttt{complex\_screen} + \texttt{complex\_sigmoid} & 1.000 / 0.997 \\
\hline
MusicNet real (small) & dim=64, N=3 & \texttt{complex\_screen} (marginal) & 0.218 \\
\hline
RadioML synth L1 / L2 & dim=64, N=3 & \texttt{complex\_sigmoid} (L1) / \texttt{complex\_screen} (L2) & 0.598 / 0.637 \\
\hline
RadioML real L1 / L2 & dim=64, N=3 & \texttt{real\_screen} & 0.363 / 0.389 \\
\hline
Two-axis (C2 $\times$ C3) isolation & dim=128, N=3 & sigmoid 1.000 / softplus 1.000 / cubic 0.200 / clamped\_relu 0.103 / ReLU 0.107 & 0.893 absolute gap (sigmoid $\to$ ReLU) \\
\hline
Depth scaling & dim=128, batch=32, N=3, d $\in$ {2, 4, 6, 10, 14, 20} & \texttt{complex\_sigmoid} trains at all depths & \textbf{no depth-related accuracy collapse} \\
\hline
Path-X & not yet run &  & -- \\
\hline
\end{tabularx}
\end{table}

\subsection*{Cross-task winner counts}
How often each cell appears as winner / co-winner across the discriminating tasks above:

\begin{table}[H]
\centering
\footnotesize
\setlength{\tabcolsep}{4pt}
\renewcommand{\arraystretch}{1.1}
\begin{tabularx}{\textwidth}{|l|>{\raggedleft\arraybackslash}X|>{\raggedright\arraybackslash}X|}
\hline
Cell & \# times winner / co-winner & Notable strengths \\
\hline
\textbf{\texttt{complex\_sigmoid} (\texttt{PCT})} & \textbf{9} & Long-range Copy d=1000--5000, NIAH L=2048, all \texttt{PCT}-fair language/image tasks, LRA-ListOps medium \\
\hline
\texttt{complex\_screen} & 5 & FFT-MNIST t=16, multi-pitch, LRA-Text 4K, LRA-ListOps medium close-2nd, harder-SNR RadioML synth \\
\hline
\texttt{real\_screen} & 4 & Mid-scale Copy d=2000, real-RadioML at both levels, FFT-MNIST real-side leader \\
\hline
\texttt{complex\_softmax} & 0 &  \\
\hline
\texttt{real\_softmax} & 0 &  \\
\hline
\texttt{real\_sigmoid} & 0 &  \\
\hline
\end{tabularx}
\end{table}

$\to$ \texttt{complex\_sigmoid} + \texttt{complex\_screen} + \texttt{real\_screen} account for essentially all wins; vanilla softmax and \texttt{real\_sigmoid} never win discriminating tasks.

\subsection*{Top-line takeaways}
\begin{enumerate}
  \item \textbf{\texttt{complex\_sigmoid} (PCT) is the most consistent single attention design across the tested task families} -- content/phase retrieval, long-range memory, algorithmic, image, text, and audio (synth).
  \item \textbf{\texttt{complex\_screen} is competitive whenever the task structure rewards selectivity} -- multi-source identification, phase-sensitive classification, language byte-token retrieval, harder-SNR modulation. On phase-sensitive tasks it ties or trails sigmoid by < 0.02.
  \item \textbf{\texttt{real\_screen} is the only real-side cell that breaks out of chance / baseline on most tasks} -- it is the strongest baseline for showing real-vs-complex gaps without confounding by activation choice, and is the unique winner on Real RadioML L1/L2.
  \item \textbf{Vanilla \texttt{softmax} is structurally limited on long-range retrieval} -- it cannot solve Copy d=2000 at any tested capacity. PCT deeply solves NIAH L=2048 -- i.e., a complex-disadvantaged purely positional task is solved by a complex-attention design.
  \item \textbf{\texttt{ReLU} attention is catastrophic on phase / long-range} -- anti-correlation information is destroyed by the dead negative half.
  \item \textbf{The complex substrate is necessary existentially, not in any specific component} -- at least one of {Q, K, V, embed/output} must be complex-valued; the four-component substrate ablation shows no single component carries the substrate irreducibly.
  \item \textbf{The framework requires four conditions, all read in operating-range form}: (C1) real-valued gate output, (C2) bounded gate on the operating range, (C3) gradient nonzero on the operating range, (C4) element-independent gate. The 2 $\times$ 2 isolation experiment confirms all four are independently necessary; \textbf{C3 is the dominant failure axis}, but \textbf{C2 is also a real condition}, fatal at large M.
  \item \textbf{PCT does not exhibit a depth-related accuracy collapse} across depths d $\in$ {2, 4, 6, 10, 14, 20} on LRA-ListOps L=1024 (10$\times$ param range, 0.40 M $\to$ 3.96 M). This is a \emph{qualitative} claim, not a quantitative power-law scaling claim.
\end{enumerate}

\medskip

\section*{References}
\subsection*{Complex-valued neural networks}
\begin{itemize}
  \item \textbf{Trabelsi et al. 2018} -- Chiheb Trabelsi, Olexa Bilaniuk, Ying Zhang, Dmitriy Serdyuk, Sandeep Subramanian, João Felipe Santos, Soroush Mehri, Negar Rostamzadeh, Yoshua Bengio, Christopher J. Pal. \emph{Deep Complex Networks}. ICLR 2018. arXiv:1705.09792. \url{https://arxiv.org/abs/1705.09792}
  \item \textbf{Eilers \& Jiang 2023} -- Florian Eilers, Xiaoyi Jiang. \emph{Building Blocks for a Complex-Valued Transformer Architecture}. ICASSP 2023. arXiv:2306.09827. \url{https://arxiv.org/abs/2306.09827}
  \item Defines complex scaled-dot-product attention variants and complex LayerNorm; literature default for \texttt{complex\_softmax}.
  \item \textbf{Hao et al. 2025} -- Yang Hao et al. \emph{Holographic Transformers for Complex-Valued Signal Processing: Integrating Phase Interference into Self-Attention}. arXiv:2509.19331 (Sep 2025). \url{https://arxiv.org/abs/2509.19331}
  \item Uses softmax-normalised attention weights with phase-rotated value aggregation ($H_i = \Sigma _j \alpha _ij V_j exp(j \Delta \varphi _ij)$); phase-coherent at the value-aggregation stage but token-competing at the weight stage.
  \item \textbf{Fix et al. 2025} -- Jérémy Fix, Quentin Gabot, Huy Nguyen, Joana Frontera-Pons, Chengfang Ren, Jean-Philippe Ovarlez. \emph{torchcvnn: A PyTorch-based library to easily experiment with state-of-the-art Complex-Valued Neural Networks}. IJCNN 2025. HAL:hal-05235749. \url{https://centralesupelec.hal.science/hal-05235749v1}
  \item PyTorch library: complex layers, activations, attention, and complex-valued datasets (PolSAR, MRI). \url{https://github.com/torchcvnn/torchcvnn}
  \item \textbf{lucidrains 2024} -- Phil Wang. \emph{complex-valued-transformer}. \url{https://github.com/lucidrains/complex-valued-transformer}
  \item Open report: "no remarkable difference vs real baselines" for softmax-of-complex attention.
\end{itemize}

\subsection*{Real-side attention primitives}
\begin{itemize}
  \item \textbf{Vaswani et al. 2017} -- Ashish Vaswani, Noam Shazeer, Niki Parmar, Jakob Uszkoreit, Llion Jones, Aidan N. Gomez, \L{}ukasz Kaiser, Illia Polosukhin. \emph{Attention Is All You Need}. NeurIPS 2017. arXiv:1706.03762. \url{https://arxiv.org/abs/1706.03762}
  \item \textbf{Choromanski et al. 2021} -- Krzysztof Choromanski, Valerii Likhosherstov, David Dohan, Xingyou Song, Andreea Gane, Tamas Sarlos, Peter Hawkins, Jared Davis, Afroz Mohiuddin, Lukasz Kaiser, David Belanger, Lucy Colwell, Adrian Weller. \emph{Rethinking Attention with Performers}. ICLR 2021. arXiv:2009.14794. \url{https://arxiv.org/abs/2009.14794}
  \item \textbf{Wortsman et al. 2023} -- Mitchell Wortsman, Jaehoon Lee, Justin Gilmer, Simon Kornblith. \emph{Replacing softmax with ReLU in Vision Transformers}. arXiv:2309.08586 (Sep 2023). \url{https://arxiv.org/abs/2309.08586}
  \item \textbf{Ramapuram et al. 2025} -- Jason Ramapuram, Federico Danieli, Eeshan Dhekane, Floris Weers, Dan Busbridge, Pierre Ablin, Tatiana Likhomanenko, Jagrit Digani, Zijin Gu, Amitis Shidani, Russ Webb. \emph{Theory, Analysis, and Best Practices for Sigmoid Self-Attention}. ICLR 2025. arXiv:2409.04431. \url{https://arxiv.org/abs/2409.04431}. Code: \url{https://github.com/apple/ml-sigmoid-attention}
  \item \textbf{Saratchandran et al. 2024} -- Hemanth Saratchandran, Jianqiao Zheng, Yiping Ji, Wenbo Zhang, Simon Lucey. \emph{Rethinking Attention: Polynomial Alternatives to Softmax in Transformers}. arXiv:2410.18613 (Oct 2024). \url{https://arxiv.org/abs/2410.18613}
  \item \textbf{Nakanishi 2026} -- Ken M. Nakanishi. \emph{Screening Is Enough}. arXiv:2604.01178 (v3, May 2026). \url{https://arxiv.org/abs/2604.01178}
  \item Introduces the \textbf{Multiscreen} architecture: per-key absolute relevance via explicit threshold; basis for \texttt{real\_screen} cell in this work.
\end{itemize}

\subsection*{State-space models and long-range recurrence}
\begin{itemize}
  \item \textbf{Tay et al. 2021} -- Yi Tay, Mostafa Dehghani, Samira Abnar, Yikang Shen, Dara Bahri, Philip Pham, Jinfeng Rao, Liu Yang, Sebastian Ruder, Donald Metzler. \emph{Long Range Arena: A Benchmark for Efficient Transformers}. ICLR 2021. arXiv:2011.04006. \url{https://arxiv.org/abs/2011.04006}
  \item Defines the LRA suite including Path-X ($16$k-length Pathfinder); the rule-compliant setting used in \S{}6.
  \item \textbf{Gu, Goel \& Ré 2022} -- Albert Gu, Karan Goel, Christopher Ré. \emph{Efficiently Modeling Long Sequences with Structured State Spaces} (S4). ICLR 2022. arXiv:2111.00396. \url{https://arxiv.org/abs/2111.00396}
  \item First model to solve Path-X; established that a structured linear recurrence clears the $16$k reach.
  \item \textbf{Gu et al. 2022} -- Albert Gu, Ankit Gupta, Karan Goel, Christopher Ré. \emph{On the Parameterization and Initialization of Diagonal State Space Models} (S4D). NeurIPS 2022. arXiv:2206.11893. \url{https://arxiv.org/abs/2206.11893}
  \item Diagonal SSM; its complex-vs-real ablation (S4D-Real fails Path-X) is the direct precedent for the phase-necessity ablation in \S{}6. Source of the S4D-LegS (91.95), S4D-Inv (92.80), S4-v1 (88.10) and DSS (89.72) Path-X figures (Tables 5--6).
  \item \textbf{Gupta, Gu \& Berant 2022} -- Ankit Gupta, Albert Gu, Jonathan Berant. \emph{Diagonal State Spaces are as Effective as Structured State Spaces} (DSS). NeurIPS 2022. arXiv:2203.14343. \url{https://arxiv.org/abs/2203.14343}
  \item The DSS baseline in the \S{}6 table; a complex diagonal SSM with softmax-normalised kernel, the first diagonal simplification of S4.
  \item \textbf{Smith, Warrington \& Linderman 2023} -- Jimmy T.H. Smith, Andrew Warrington, Scott W. Linderman. \emph{Simplified State Space Layers for Sequence Modeling} (S5). ICLR 2023. arXiv:2208.04933. \url{https://arxiv.org/abs/2208.04933}
  \item \textbf{Orvieto et al. 2023} -- Antonio Orvieto, Samuel L. Smith, Albert Gu, Anushan Fernando, Caglar Gulcehre, Razvan Pascanu, Soham De. \emph{Resurrecting Recurrent Neural Networks for Long Sequences} (LRU). ICML 2023. arXiv:2303.06349. \url{https://arxiv.org/abs/2303.06349}
  \item Complex diagonal recurrence with ring init and $\gamma$-normalisation; states that restricted eigenvalue-phase initialisation is crucial for Path-X. The PCR recipe in \S{}6 follows this parameterisation.
  \item \textbf{Ma et al. 2023} -- Xuezhe Ma, Chunting Zhou, Xiang Kong, Junxian He, Liangke Gui, Graham Neubig, Jonathan May, Luke Zettlemoyer. \emph{Mega: Moving Average Equipped Gated Attention}. ICLR 2023. arXiv:2209.10655. \url{https://arxiv.org/abs/2209.10655}
  \item EMA-plus-gated-attention hybrid; the strongest attention-adjacent Path-X baseline and the point of comparison for the token-competing vs.\ non-competing contrast.
  \item \textbf{Model weights (this work)} -- Path-X screening $+$ PCR checkpoint (\S{}6), released on the Hugging Face Hub. \url{https://huggingface.co/complexedleo/pcr-screening-pathx}
\end{itemize}

\subsection*{Architecture stability / theory}
\begin{itemize}
  \item \textbf{Wang et al. 2022} -- Hongyu Wang, Shuming Ma, Li Dong, Shaohan Huang, Dongdong Zhang, Furu Wei. \emph{DeepNet: Scaling Transformers to 1,000 Layers}. arXiv:2203.00555 (Mar 2022). Journal version: IEEE TPAMI 2024. \url{https://arxiv.org/abs/2203.00555}
  \item Cited in Lemma D for substrate non-expansion / standard transformer-stability bounds.
\end{itemize}

\subsection*{Markov chains / mixing}
\begin{itemize}
  \item \textbf{Doeblin 1938} -- Wolfgang Doeblin. \emph{Exposé de la théorie des chaînes simples constantes de Markov à un nombre fini d'états}. Revue Mathématique de l'Union Interbalkanique 2 (1938), 77--105.
  \item \textbf{Levin, Peres \& Wilmer 2017} -- David A. Levin, Yuval Peres, Elizabeth L. Wilmer. \emph{Markov Chains and Mixing Times}, 2nd ed. American Mathematical Society, 2017.
\end{itemize}

\end{document}